\RequirePackage{fix-cm}
\documentclass[twocolumn]{svjour3}          
\smartqed  
\usepackage{graphicx}
\usepackage{balance}       
\usepackage{graphics}      
\usepackage{color}
\usepackage{booktabs}

\usepackage{graphicx}

\usepackage{adjustbox}
\usepackage{caption,subcaption,graphicx}

\usepackage{tikz}
\def\checkmark{\tikz\fill[scale=0.4](0,.35) -- (.25,0) -- (1,.7) -- (.25,.15) -- cycle;}
\usepackage{multirow}
\usepackage{graphicx}

\usepackage{enumitem}
\usepackage{comment}
\usepackage{seqsplit}

\usepackage{hyphenat}
\usepackage{xstring}
\usepackage{forloop}
\usepackage{soul}

\newsavebox\MyBreakChar%
\sbox\MyBreakChar{\hyp}
\newsavebox\MySpaceBreakChar%
\sbox\MySpaceBreakChar{}
\makeatletter%
\newcommand*{\BreakableChar}[1][\MyBreakChar]{%
  \leavevmode%
  \prw@zbreak%
  \discretionary{\usebox#1}{}{}%
  \prw@zbreak%
}%

\newcounter{index}%
\newcommand{\AddBreakableChars}[1]{%
  \StrLen{#1 }[\stringLength]%
  \forloop[1]{index}{1}{\value{index}<\stringLength}{%
    \StrChar{#1}{\value{index}}[\currentLetter]%
    \IfStrEq{\currentLetter}{ }
        {\currentLetter\BreakableChar[\MySpaceBreakChar]}%
        {\currentLetter\BreakableChar[\MyBreakChar]}%
  }%
}%

\colorlet{RED}{red}

\begin{document}

\title{Enabling AI and Robotic Coaches for Physical Rehabilitation Therapy
}
\subtitle{{Iterative Design and Evaluation with Therapists and Post-Stroke Survivors}}



\author{Min Hun Lee         \and
        Daniel P. Siewiorek \and
        Asim Smailagic \and
        Alexandre Bernardino \and
        Sergi Bermúdez i Badia
}


\institute{Min Hun Lee \at Singapore Management University\\
            \email{mhlee@smu.edu.sg} 
            \and
           Daniel P. Siewiorek and Asim Smailagic \at
              Carnegie Mellon University \\
              \email{\{dps, asim\}@cs.cmu.edu}           
           \and
           Alexandre Bernardino \at
           Instituto Superior Técnico\\
              \email{alex@isr.tecnico.ulisboa.pt}           
           \and
           Sergi Bermúdez i Badia \at
           University of Madeira, NOVA-LINCS\\
              \email{sergi.bermudez@staff.uma.pt}           
}

\date{Published at International Journal of Social Robotics in 2022; https://doi.org/10.1007/s12369-022-00883-0}

\maketitle

\begin{abstract}
Artificial intelligence (AI) and robotic coaches promise the improved engagement of patients on rehabilitation exercises through social interaction. While previous work explored the potential of automatically monitoring exercises for AI and robotic coaches, the deployment of these systems remains a challenge. Previous work described the lack of involving stakeholders to design such functionalities as one of the major causes. In this paper, we present our efforts on eliciting the detailed design specifications on how AI and robotic coaches could interact with and guide patient's exercises in an effective and acceptable way with four therapists and five post-stroke survivors. Through iterative questionnaires and interviews, we found that both post-stroke survivors and therapists appreciated the potential benefits of AI and robotic coaches to achieve more systematic management and improve their self-efficacy and motivation on rehabilitation therapy. In addition, our evaluation sheds light on several practical concerns (e.g. a possible difficulty with the interaction for people with cognitive impairment, system failures, etc.). We discuss the value of early involvement of stakeholders and interactive techniques that complement system failures, but also support a personalized therapy session for the better deployment of AI and robotic exercise coaches.

\end{abstract}

\maketitle
\section{Introduction}
Physical rehabilitation therapy is one of the effective approaches to improve the functional abilities of patients with neurological and musculoskeletal problems \cite{van2004impact}. As therapists are not always able to monitor and guide patient's repetitive rehabilitation therapy, they often prescribe self-directed exercises \cite{peek2016interventions}. An outcome of physical rehabilitation therapy highly depends on how much a patient adheres to perform prescribed rehabilitation exercises \cite{thomason2013rehabilitation}. However, the adherence to repetitive rehabilitation therapy over an extended period is challenging for patients without the presence of a therapist \cite{thomason2013rehabilitation}.  Patient's low adherence to the prescribed, self-directed exercises is a common problem across several healthcare disciplines of physiotherapy \cite{argent2018patient}. 

To address this problem, there has been increasing attention on artificial intelligence (AI) and robotic coaches \cite{van2016rehabilitation}. {These systems leverage AI techniques to autonomously monitor patient's exercises \cite{lee2020exploratory}. In addition, these systems can assist patient's engagement in well-being-related or rehabilitation exercises through social interaction (e.g. providing encouragement \cite{fasola2013socially,mataric2007socially,lee2020towards}). In this paper, we focus on a system that leverages AI techniques and robotic embodiment to coach exercises and refer it as an AI and robotic coach.}

Researchers have focused on addressing the technical challenges of automatically monitoring patient's exercises using sensors and machine learning \cite{lee2019learning,tanguy2016computational}. In addition, prior work has evaluated AI and robotic exercise coaches with a few design variables (e.g. the effect of physical embodiment \cite{fasola2013socially}, matching the style of interaction with user's personality \cite{tapus2008user}), and shown a positive impact on patient's engagement and motivation \cite{swift2015effects,gockley2006encouraging}. However, even if prior work has demonstrated the feasibility of specific functionalities of AI and robotic exercise coaches, it is still challenging to adopt these systems broadly \cite{pripfl2016results,winkle2018social,wilson2020challenges}. These challenges include safety, clinical effectiveness, cost-effectiveness, usability \cite{riek2017healthcare}. In this paper, we primarily focus on the usability aspect. We build upon prior work that discusses the need of involving the target stakeholders in an early design phase \cite{winkle2018social,wilson2020challenges} to understand their perspectives on the real-world context and design AI and robotic coaches more acceptable in practice \cite{baillie2019challenges}.

In this work, we describe detailed design specifications and exploratory evaluations on an AI and robotic coach that can monitor and guide post-stroke survivor's self-paced physical rehabilitation therapy. Based on findings from interviews with therapists and post-stroke survivors, we designed and developed an AI and robotic exercise coach with six main functionalities: 1) planning, 2) initiating, 3) introducing a session, 4) monitoring and providing corrective feedback, 5) adapting the difficulty of a session, and 6) concluding a session. After developing a system, we evaluated the potential of this system with therapists and post-stroke survivors through questionnaires and interviews before and after showing video demonstrations of the system. 

Overall, both therapists and patients expressed the potential benefits of our system to support more systematic management of self-paced rehabilitation and improve the self-efficacy and motivation of post-stroke survivors. In addition, the findings revealed several practical concerns on using an AI and robotic coach: probable difficulty with the interaction for people with cognitive impairment, diversified ways to interact with a system, strategies to manage system failures, portability, and cost-efficiency. While addressing these concerns in the near future is challenging, we discuss the importance of early involvement of stakeholders and interactive techniques that have the potential to address system failures and support a personalized session for deploying AI and robotic exercise coaches. To our knowledge, this work is the first to design and elicit opinions on {the detailed design specifications of} an AI and robotic exercise coach from both therapists and post-stroke survivors {for} the overall process of a self-directed, post-stroke rehabilitation session (e.g. planning, initiating, introducing, monitoring, adapting, and concluding a session).

\section{Related Work}\label{sect:related}
In this section, we describe the background on rehabilitation for post-stroke survivors and outline related work of technological supports on rehabilitation: {motion tracking technologies and applications for patients to improve their engagement in exercises including efforts on human-centered designs of social robots.}

\subsection{Practices of Post-Stroke Physical Rehabilitation}
A stroke occurs when the blood supply to the brain is interrupted or reduced and brain cells die. Such an injury on brain cells has a significant impairment on cognitive and motor abilities of post-stroke survivors \cite{ekstam2007functioning}. Post-stroke survivors require rehabilitation interventions over an extended period to improve their function and independence in daily activities \cite{o2019physical}. During rehabilitation interventions, therapists assess the condition of a patient using various methods (e.g. reviewing patient's history \cite{o2019physical} or conducting clinical tests that require the therapist's direct observation of the patient's motions 
\cite{sullivan2011fugl}) and discuss with a patient to set a goal for improvement. Performing a task-oriented exercise (e.g. bring a cup to the mouth) is one of the effective interventions \cite{rensink2009task} to regain the patient's functional ability. After interventions, therapists re-assess the patient's progress and modify interventions if necessary \cite{o2019physical}. 

As therapists have limited availability to support repetitive rehabilitation sessions from post-stroke survivors \cite{long2003rehabilitation}, therapists rely on prescribing self-directed exercises in-between therapy sessions \cite{peek2016interventions}. However, post-stroke survivors have low engagement with their self-directed rehabilitation without a therapist's supervision \cite{argent2018patient}.

\subsection{Technological Support for Physical Rehabilitation}\label{sect:related-tech}
Researchers have investigated various technologies to facilitate the delivery of physical rehabilitation \cite{loureiro2011advances}. 

\subsubsection{Motion Tracking Techniques}\label{sect:hri-study-design-related-motion-track}
One fundamental technology of rehabilitation is a motion tracking system that dynamically represents the pose of a human body using sensors. These motion tracking systems can be categorized into non-visual sensors (e.g. inertial, magnetic, etc.) and visual marker-based or marker-free \cite{zhou2008human}. Among various approaches, a visual marker-based technique that leverages infrared cameras capturing motions from reflective markers on the human body is often considered as a golden standard due to their highest performance (i.e. errors around 1mm) \cite{zhou2008human}. However, it has a limitation due to its complex set-up that requires an expert operation and expensive costs \cite{do2016movement}. In contrast, both non-visual, inertia sensors and visual marker-free systems provide competitive performance for rehabilitation monitoring \cite{zhou2008human} and lower cost for patients and clinicians \cite{do2016movement}. As inertia sensors have limitations of measurement due to inconsistent positions of sensors and cumbersome wear sensors, this work applies a visual marker-free technique (i.e. a Kinect sensor) to track patient's rehabilitation exercises.

\subsubsection{Applications for Patients}\label{sect:hri-study-design-related-app-patients}
Motion tracking techniques can be further developed into various applications for better rehabilitation experiences for a patient.
These applications include virtual reality \cite{subramanian2007virtual}, intelligent coaching systems \cite{lee2019learning}, and assistive robots \cite{lo2012exoskeleton,fasola2013socially,lee2020towards}. Building upon a motion tracking technique with sensors, these systems aim to provide engaging experiences or richer information on rehabilitation. For instance, researchers have utilized computer-simulated interactive environments \cite{subramanian2007virtual} or games \cite{alankus2010towards} to promote patient's participation in rehabilitation. Exoskeleton robots have been {explored} to {augment patient's weak body limbs} and induce a passive motion {for rehabilitation} \cite{lo2012exoskeleton}. AI \cite{lee2019learning} or robotic coaches \cite{fasola2013socially,van2016rehabilitation,lee2020towards} can guide patient's rehabilitation through automatically monitoring patient's exercises \cite{lee2020exploratory} and providing feedback on {whether a patient performs} well-being-related or rehabilitation exercises {correctly or not} \cite{fasola2013socially,mataric2007socially,swift2015effects,gockley2006encouraging,lee2020towards}. As prior work has demonstrated the benefit of physical embodiment to improve the engagement in physical exercises \cite{fasola2013socially}, we decided to further explore research on socially assistive robotics.

A large body of work on socially assistive robotics has focused on a specific technical improvement (e.g. improving a technique of automated assessment \cite{tanguy2016computational,lee2020towards}) or the effect of a particular design variable (e.g. physical embodiment \cite{fasola2013socially}, matching the style of interaction with user's personality \cite{tapus2008user}, the usage of comparative feedback \cite{swift2015effects}). Prior work has shown the potential of a socially assistive robot to improve patient's engagement in well-being related or rehabilitation exercises \cite{fasola2013socially,swift2015effects,tapus2008user}. However, prior work does not explore the entire pipeline of a rehabilitation session (e.g. from planning to conducting a session) and assumes that the end-user will initiate interaction with a system. In addition, no solutions have been widely adopted \cite{pripfl2016results,winkle2018social}.

\subsection{Human Centered Designs of Social Robots}
For better real-world deployment of socially assistive robots, researchers have employed user-centered design and evaluation methods to elicit user needs and derive design requirements \cite{beer2012domesticated,azenkot2016enabling,winkle2018social,lee2021interactive}. Beer et al. utilized narrated videos of the robot to conduct the needs assessment of elderly people on assistive robots through questionnaires and structured group interviews \cite{beer2012domesticated}. They provided preliminary recommendations of mobile manipulator robots to support aging in place \cite{beer2012domesticated}. Azenkot et al. derived design specifications of building service robots that guide blind people in a large building through multiple sessions (e.g. interviews and a group workshop) between designers and blind people  \cite{azenkot2016enabling}.
Winkle et al. described design guidelines of social robots for rehabilitation, from focus group sessions and interviews with therapists \cite{winkle2018social}. In addition, Polak and Levy-Tzedek also conducted focus group sessions with therapists and a preliminary evaluation study on a gamification system for rehabilitation with four post-stroke survivors \cite{feingold2020social}. Although both \cite{winkle2018social} and \cite{feingold2020social} provide design recommendations, they do not incorporate the opinions of the end-user (e.g. post-stroke survivors), who will interact with the system. It remains unclear about the detailed design specifications on how AI and robotic coaches could interact with and guide patient's rehabilitation. 

While there has been a lot of research on applications for patients, specifically social robots for rehabilitation, our work differs in two key aspects. First, we involved both therapists and post-stroke survivors to understand their practices and needs and seek to design how an AI and robotic coach could interact with and guide post-stroke survivor's self-directed rehabilitation. Prior work described studies with therapists to derive design recommendations \cite{winkle2018social,feingold2020social}, but both \cite{winkle2018social} and \cite{feingold2020social} do not involve the end-user (i.e. post-stroke survivors) in their design processes. In addition, we conducted additional interviews with therapists and post-stroke survivors to understand their opinions about an AI and robotic coach during the overall process of self-directed rehabilitation \cite{lee2021interactive} instead of focusing on only an individual step of rehabilitation therapy (e.g. monitoring an exercise \cite{fasola2013socially,swift2015effects,tapus2008user}).

\section{Study on an AI and Robotic Coach for Physical Stroke Rehabilitation Therapy}
In this work, we aim to explore the potential of an AI and robotic coach for physical stroke rehabilitation therapy. Specifically, this research aims to 1) understand the needs of post-stroke survivors during self-directed rehabilitation and the practices of therapists to guide a rehabilitation session, 2) seek design specifications that detail how an AI and robotic coach can interact with and assist post-stroke survivor's self-directed rehabilitation, and 3) understand opinions of therapists and post-stroke survivors to use this system. Based on the findings of our study, we discuss the implications to design an AI and robotic coach.

\subsection{Research Team and Participants}

We created an interdisciplinary team made up of three human-computer interaction (HCI) researchers, one \AddBreakableChars{robotics} researcher, and one neurorehabilitation researcher. We then recruited four therapists with experiences of stroke rehabilitation (Table \ref{tab:hri-study-design-tps}) and five post-stroke survivors (Table \ref{tab:hri-study-design-patients}) through email communication to local hospitals and contacts of the research team.
Both therapists and post-stroke survivors were involved throughout the study for human-centered design and evaluation of AI and robotic coaches. 

To collect diverse opinions during interviews, we recruited four therapists from three rehabilitation centers (1 male and 3 females; 35.75 $\pm$ 7.14 years old) with various experiences and disciplines: $\mu=12.50$, $\sigma=9.04$ years in stroke rehabilitation; 3 occupational therapists, who focus on helping patients to better engage in their daily livings and 1 physiotherapist, who treats patient's actual impairment from a biomechanical perspective (Table \ref{tab:hri-study-design-tps}). The occupational therapists (TP 1, TP 2, TP 3) have also experience as physiotherapists during their careers. In addition, we ensured the diversity of five post-stroke survivors (4 males and 1 female; 59.00 $\pm$ 4.64 years old) with various phases of the stroke, functional abilities, and experiences in stroke rehabilitation: $\mu=3.16$, $\sigma=3.59$ years since stroke; one post-stroke with low functional ability without voluntary control of hands (PS 2), three moderate functional abilities (PS 1, 3, 5), and one high functional ability (PS 4). 
We also collected post-stroke survivors' experience with technology through questionnaires measuring familiarity with technologies designed by the Center for Research and Education on Aging and Technology Enhancement (CREATE) \cite{beer2012domesticated,czaja2006factors}. Post-stroke survivors rated their experience with technologies on a 7-point scale (1 = strongly disagree, 2 = disagree, 3 = slightly disagree, 4 = neutral, 5 = slightly agree, 6 = agree, 7 = strong agree on experience with technology, personal computers, smartphones, and robots). A low score on technology experience (e.g. 1.0) indicates that a post-stroke survivor barely has experience with recent technologies (e.g. a personal computer, a smartphone, a robot).

Overall, post-stroke survivors have diverse levels of experience with technology (3.20 $\pm$ 2.05 score of technology experience in Table \ref{tab:hri-study-design-patients}).
Three post-stroke survivors reported that they are somewhat familiar with technology, personal computers, and smartphones, but two post-stroke survivors with 1.4 and 1.0 technology experience scores reported no experience with smartphones. None of the post-stroke survivors have experience with robots. 

\begin{table*}[htp]
\centering
\caption{Profiles of Therapists on Initial Interview, Review, and Evaluation on the System}
\label{tab:hri-study-design-tps}
\resizebox{1.0\columnwidth}{!}{%
\begin{tabular}{@{}ccccc@{}}
\toprule
ID &
  \begin{tabular}[c]{@{}c@{}}Interview\\ (P1-a)\end{tabular} &
  \begin{tabular}[c]{@{}c@{}}Review\\ (P2-b)\end{tabular} &
  \begin{tabular}[c]{@{}c@{}}Evaluation\\ (P3-a)\end{tabular} &
  \begin{tabular}[c]{@{}c@{}}\# of Years in\\ Stroke Rehabilitation\end{tabular} \\ \midrule
TP 1 & \checkmark & \checkmark &   & 6  \\
TP 2 & \checkmark & \checkmark &   & 4  \\
TP 3 & \checkmark &   & \checkmark & 23 \\
TP 4 &   &   & \checkmark & 17 \\ \bottomrule
\end{tabular}%
}
\end{table*}

\begin{table*}[htp]
\centering
\caption{Profiles of Post-Stroke Survivors on Initial Interview and Evaluation on the System}
\label{tab:hri-study-design-patients}
\resizebox{0.8\textwidth}{!}{%
\begin{tabular}{@{}ccccccc@{}}
\toprule
\textbf{ID} &
  \textbf{Sex} &
  \textbf{Age} &
  \begin{tabular}[c]{@{}c@{}}\textbf{Type of} \\ \textbf{Stroke}\end{tabular} &
  \begin{tabular}[c]{@{}c@{}}\textbf{\# of Years}\\ \textbf{since Stroke}\end{tabular} &
  \begin{tabular}[c]{@{}c@{}}\textbf{Functional}\\ \textbf{Ability/Status}\end{tabular} &
  \begin{tabular}[c]{@{}c@{}}\textbf{Technology}\\ \textbf{Experience }\cite{czaja2006factors}\end{tabular} \\ \midrule
PS 1 & Male   & 61 & Ischemic & 8.0 years & \begin{tabular}[c]{@{}c@{}}Moderate\\ FMA score: 47 of 66\end{tabular}                      & 5.8 of 7.0 \\ \midrule
PS 2 &
  Male &
  54 &
  Hemorrhagic &
  \begin{tabular}[c]{@{}c@{}}0.7 years\end{tabular} &
  \begin{tabular}[c]{@{}c@{}}Low; No voluntary control of hands\\ FMA score: N/A\end{tabular} &
  3.2 of 7.0 \\ \midrule
PS 3 & Female & 65 & Ischemic & 1.0 years & \begin{tabular}[c]{@{}c@{}}Moderate\\ FMA score: 36 of 66\end{tabular}                      & 1.4 of 7.0 \\ \midrule
PS 4 & Male   & 59 & Ischemic & 6.0 years & \begin{tabular}[c]{@{}c@{}}High\\ FMA score: 66 of 66\\ light hearing problems\end{tabular} & 4.6 of 7.0 \\ \midrule
PS 5 &
  Male & 55
   &
  Hemorrhagic &
  \begin{tabular}[c]{@{}c@{}}Initial: 1.6 years\\ Relapse: 0.5 years\end{tabular} &
  \begin{tabular}[c]{@{}c@{}}Moderate\\ FMA score: N/A\\ Needs supervision on daily living activites\end{tabular} & 1.0 of 7.0
   \\ \bottomrule
\end{tabular}%
}
\end{table*}

\subsection{Procedure}
This study consisted of a series of interviews with four therapists with experience in stroke rehabilitation and five post-stroke survivors {to design and conduct the exploratory evaluation} of an AI and robotic coach for stroke rehabilitation, and the development of a {prototype} by the research team (Table \ref{tab:hri-study-design-procedures}): 1) initial interviews with five post-stroke survivors (P1-b in Table  \ref{tab:hri-study-design-procedures}) and three therapists (P1-a in Table  \ref{tab:hri-study-design-procedures}) to gain deeper understanding of their needs and practices, 2) design and development of a high-fidelity prototype from the research team (P2-a in Table \ref{tab:hri-study-design-procedures}) and interviews with each of two therapists to review the prototype before evaluation (P2-b in Table \ref{tab:hri-study-design-procedures}), and 3) interviews with therapists (P3-a in Table \ref{tab:hri-study-design-procedures}) and five post-stroke survivors (P3-b in Table  \ref{tab:hri-study-design-procedures}) to evaluate the prototype before/after showing videos of the prototype. The study procedures were approved by the institutional review board (IRB). The detailed procedures of each process are described as follows:

\begin{table*}[htp]
\centering
\caption{Overall Procedures of Our Study with Therapists and Post-Stroke Survivors}
\label{tab:hri-study-design-procedures}
\resizebox{\textwidth}{!}{%
\begin{tabular}{clcl} \toprule
\textbf{Process} &
  \multicolumn{1}{c}{\textbf{Purpose}} &
  \textbf{Participants} &
  \multicolumn{1}{c}{\textbf{Methods}} \\ \midrule
P1-a &
  Understand the challenges of post-stroke survivors during self-directed rehabilitation &
  \begin{tabular}[c]{@{}c@{}}5 Post-Stroke Survivors\\ (PS 1, 2, 3, 4, 5)\end{tabular} &
  Semi-structured Interviews \\ \midrule
P1-b &
  Learn the practices \& strategies of therapists to guide a rehabilitation session &
  \begin{tabular}[c]{@{}c@{}}3 Therapists\\ (TP 1, 2, 3)\end{tabular} &
  \begin{tabular}[c]{@{}l@{}}Semi-structured interviews \end{tabular} \\ \midrule
P2-a &
  Design and develop a high-fidelity prototype &
  Researchers & \begin{tabular}[c]{@{}l@{}}Analysis \&\\ High-Fidelity Prototyping\end{tabular}\\ \midrule
P2-b &
  Review the videos of the prototype &
  \begin{tabular}[c]{@{}c@{}}2 Therapists\\ (TP 1 \& 2)\end{tabular} &
  \begin{tabular}[c]{@{}l@{}}Interview\\ on the prototype\end{tabular} \\ \midrule
P3-a &
  \multirow{2}{*}{Understand the opinions of using a system before/after showing a video} &
  \begin{tabular}[c]{@{}c@{}}2 Therapists\\ (TP 3 \& 4)\end{tabular} &
  \multirow{2}{*}{\begin{tabular}[c]{@{}l@{}}Interview \& questionnaires\\ on the prototype\end{tabular}} \\
P3-b &
   &
  \begin{tabular}[c]{@{}c@{}}5 Post-Stroke Survivors\\ (PS 1, 2, 3, 4, 5)\end{tabular} & \\ \bottomrule
\end{tabular}%
}
\end{table*}

\subsubsection{Initial Interviews with Post-Stroke Survivors (P1-a)}
The objective of initial interviews with post-stroke survivors was to understand their challenges and needs during self-directed rehabilitation and probe their initial ideas on technological supports. One HCI researcher of the team conducted a one-on-one interview with each of the five post-stroke survivors with the assistance of a therapist. Before the interview, demographics and informed consent were collected from post-stroke survivors. During the one-hour interview, the researcher asked post-stroke survivors to describe their challenges with conducting self-paced rehabilitation therapy. In addition, the researcher explained the structure and design space of the project and the assumed capabilities of technology, and asked how technology could support their challenges to probe the ideas from post-stroke survivors (e.g. \textit{``what kinds of technical support would you like to receive during self-paced rehabilitation?''}).

\subsubsection{Initial Interviews with Therapists (P1-b)}
The objective of initial interviews with therapists was to learn their practices and strategies to guide rehabilitation sessions and their initial ideas on technological supports during self-directed rehabilitation of post-stroke survivors.
One HCI researcher of the team conducted a one-on-one interview with each of three therapists (TPs with checkmarks in the interview column of Table \ref{tab:hri-study-design-tps}; $\mu = 11.00$, $\sigma=10.44$ years of experience in stroke rehabilitation). Before the interview, demographics and informed consent were collected from therapists. During the one-hour interview, the researcher asked therapists to describe their practices to manage a rehabilitation session (i.e. \textit{``how do you operate a session''}), and speak aloud their strategies and feedback that they generate during a session (i.e. \textit{``what kinds of feedback do you generate for a post-stroke survivor?''}). To assist therapists' speaking aloud process, the researcher showed them videos of post-stroke survivors, who have different functional abilities (i.e. high, moderate, low capability to achieve an exercise) and perform rehabilitation exercises. These videos were collected by the research team in a previous study on technological support to automatically monitor stroke rehabilitation exercises \cite{lee2019learning}. At the end of the interviews, the researcher asked therapists about the possibility of technological support for self-directed rehabilitation of post-stroke survivors. 

\subsubsection{Design and Development of a Prototype (P2-a and P2-b)}
In this process, the team designed and developed an AI and robotic coach that not only meets the needs of post-stroke survivors but also follows the practices and strategies of therapists. After initial interviews with post-stroke survivors and therapists, two researchers analyzed transcripts through the process described in Section \ref{sect:hri-study-design-analysis}. With the findings, the research team further discussed the specifications of an AI and robotic coach for post-stroke physical rehabilitation and developed a high-fidelity prototype for evaluation (Section \ref{sect:hri-study-design-results-prototype}). 
The functionalities of our prototype were recorded into narrated videos to show its capabilities. 
Two therapists (TPs with check marks in the \textit{`review'} column of Table \ref{tab:hri-study-design-tps};  $\mu = 5.00$, $\sigma=1.41$ years of experience in stroke rehabilitation) reviewed these videos to detect any issues to conduct an evaluation study with post-stroke survivors.

\subsubsection{Evaluation}
The objective of this process was to seek the opinions of therapists and post-stroke survivors about how technological support might be useful in practice. During the one-hour interview, we primarily focused on collecting opinions on the overall procedures of a self-directed rehabilitation session ({e.g.} planning, initiating, introducing, monitoring, adapting, and concluding a self-directed rehabilitation session).

Both therapists (TPs with checkmarks in the \textit{`evaluation'} column of Table \ref{tab:hri-study-design-tps};  $\mu = 20.00$, $\sigma=4.24$ years of experience in stroke rehabilitation) and post-stroke survivors completed the questionnaires and provided comments about their opinions on how well our AI and robotic coach can support six major functionalities for self-directed rehabilitation.
We informed therapists and post-stroke survivors to assume that technology could perform the procedure to the level of an expert, therapist. They rated their opinions on each procedure on a 7-point scale (1 = strongly disagree, 2 = disagree, 3 = slightly disagree, 4 = neutral, 5 = slightly agree, 6 = agree, 7 = strongly agree on technological support). After completing their initial responses, both therapists and post-stroke survivors watched the narrated video of the prototype (https://youtu.be/OSpMqWZXDXo) and then completed the same questionnaires. During their second responses, they also rated the questions on comprehension and usability of each procedure, the functionality of the prototype (i.e. comprehension: \textit{``The system provides understandable interaction, feature''} and usability: \textit{``The system provides useful, valuable interaction, feature''}). In addition, they provided comments on the benefits and limitations of the prototype. 

\subsection{Analysis}\label{sect:hri-study-design-analysis}
All interviews with therapists and post-stroke survivors were audio-recorded and transcribed for analysis. We then followed a deductive and inductive approach to coding transcripts \cite{gale2013using}. Specifically, initial codes were generated based on the literature review and research questions. Two researchers then independently coded transcripts with initial codes and also generated any additional codes inductively if necessary. The codes were discussed with the team and iteratively refined. 

\section{Challenges of Post-Stroke Survivors during Self-Directed Rehabilitation}\label{sect:hri-study-design-int-patients}

According to the interviews, post-stroke survivors are conscious of the importance of rehabilitation and strive to engage in self-directed rehabilitation sessions. However, they all encounter challenges to pursue self-directed rehabilitation due to several factors: low adherence due to spontaneous planning, low efficacy, uncertainty and confusion, and lack of systematic management (Table \ref{tab:hri-study-design-prototype-overview}). Each post-stroke survivor describes different attitudes and styles of planning and managing self-directed rehabilitation.

\begin{table*}[htp]
\centering
\caption{Challenges of Post-Stroke Survivors and Corresponding Functionalities of an AI and Robotic Coach}
\label{tab:hri-study-design-prototype-overview}
\resizebox{\textwidth}{!}{%
\begin{tabular}{@{}lll@{}}
\toprule
\textbf{\begin{tabular}[c]{@{}l@{}}Challenges \\  \& Needs\end{tabular}} &
  \multicolumn{2}{l}{\textbf{Functionalities of an AI and Robotic Coach}} \\ \midrule
\multirow{2}{*}{\begin{tabular}[c]{@{}l@{}}Spontaneous Planning\\ \& Low Adherence\end{tabular}} &
  F1. Planning &
  \begin{tabular}[c]{@{}l@{}}A therapist uploads prescriptions of post-stroke survivor's self-paced rehabilitation (Figure \ref{fig:hri-design-system-F1-1}) \\ A post-stroke survivor receives notification and plan the schedules of self-directed rehabilitation (Figure \ref{fig:hri-design-system-F1-2})\end{tabular} \\
 &
  F2. Initiation &
  The robot approaches the post-stroke survivor to initiate a session (Figure \ref{fig:hri-design-system-F2}) \\ \midrule
\multirow{3}{*}{\begin{tabular}[c]{@{}l@{}}Low Efficacy, \\ Uncertainty,\\ \& Confusion\end{tabular}} &
  F3. Introduction &
  \begin{tabular}[c]{@{}l@{}}The robot describes the goal of a session and\\ shows the demonstration of an exercise with gestures and a video on the display (Figure \ref{fig:hri-design-system-F3})\end{tabular} \\
 &
  \begin{tabular}[c]{@{}l@{}}F4. Monitoring\\\hspace{0.5cm}\& Feedback\end{tabular} &
  \begin{tabular}[c]{@{}l@{}}The robot monitors and assesses post-stroke survivor's exercises and \\ provide positive encouragement and corrective feedback with gestures, audios, and visualization (Figure \ref{fig:hri-design-system-F4})\end{tabular} \\
 &
  \begin{tabular}[c]{@{}l@{}}F5. Adapting\\\hspace{0.4cm} Difficulty\end{tabular} &
  \begin{tabular}[c]{@{}l@{}}The robot communicates to understand the user's status and adjust the difficulty of a session\\ if the post-stroke survivor continuously performs an exercise with compensated joints (Figure \ref{fig:hri-design-system-F5})\end{tabular} \\ \midrule
\begin{tabular}[c]{@{}l@{}}Systematic Management\\ \& Records on Progress\end{tabular} &
  F6. Concluding &
  The robot summarizes the progress of the post-stroke survivor and reminds about the next session (Figure \ref{fig:hri-design-system-F6}) \\ \bottomrule
\end{tabular}%
}
\end{table*}

\subsection{Attentive to Value of Rehabilitation, but Challenging to Maintain Motivation}
Whether post-stroke survivors have recovered and discharged or not, rehabilitation still plays a central role to improve their functional abilities.  They are attentive to the importance of engagement in rehabilitation. For instance, PS1 has engaged in rehabilitation for 8 years after stroke, is still enrolled in physiotherapy, and conducts a self-managed exercise for his better quality of life. 

\textit{``Even if I do not have sessions with therapists anymore, I am still willing to do additional sessions myself to maintain my motor skills''} (PS 4). \textit{``I do more therapy exercises at home myself. I never stop to get better even after arriving home from a weekly rehabilitation with a therapist''} (PS 2).

Post-stroke survivors require continuous engagement in rehabilitation for an extended period to improve their functional abilities \cite{o2019physical}. However, they encounter various challenges to maintain their motivation and engagement in rehabilitation: spontaneous planning of a session due to internal and external factors, low efficacy on the program and correct execution of exercises, and lack of systematic management and recording on progress. 

\textit{``Having positive recovery''} (PS 4) and \textit{``internal motivation are critical to keep rehabilitation up every day''} (PS 2). However, \textit{``Sometimes, I do not feel motivated to do any exercises.''} (PS 3). 

\subsection{Low Adherence due to Spontaneous Planning}\label{sect:hri-study-design-int-patients-planning}
Post-stroke survivors strive to engage in rehabilitation whenever they are available with the hope of improving their functional abilities. They describe different styles to plan their sessions. Some post-stroke survivors attempt to incorporate their self-paced rehabilitation into their routines and make a plan every day. Others just make mental planning every weekend or whenever they recalled and are available. Whether they make high-level mental plans or specific daily plans, they mostly end up having spontaneous planning due to various external and internal factors. For instance, a planned session is sometimes delayed or canceled due to the availability of transportation and a place to conduct a session. Depending on the feelings, physical conditions, and personal schedules, post-stroke survivors often manage their self-paced sessions spontaneously.

\textit{``I plan what I’m going to do in the morning and the afternoon. I update my internal mental plan to do a little more or a new exercise, depending on other personal schedules, feeling, or my progress - whether I could move a little bit forward''} (PS 2).

\textit{``On Sunday afternoons, I plan my week schedules of training in my mind, but these schedules are often changed. If the place is available and I have transportation, I always try to go there. If not, I just do not conduct any sessions.''} (PS 1). 

\textit{``I do not specify a time to start or finish''} (PS 5) \textit{``I work on exercises when I remember and feel like it and remember it. Sometimes, I end up forgetting about it and just remember on my bed before sleeping''} (PS 3). 

Such spontaneous planning can lead to low adherence to self-paced rehabilitation sessions, and even degrade the functional ability of a patient. \textit{``I used to perform balance exercises easily for 5 minutes, but after continuously missing my self-paced sessions during quarantine from COVID, I am not able to it at all now.''} (PS 1). 
\textit{``after skipping my daily exercises, I have difficulty with moving on the next day and feel consequences''} (PS 3).

\subsection{Low Efficacy, Uncertainty and Confusion}
Even after successful management to start a self-paced rehabilitation session, all post-stroke survivors encounter another challenge of being uncertain and confused on various aspects of a session.

\subsubsection{Program of Exercises}\label{sect:hri-study-design-int-patients-program}
At the beginning of a self-paced session, post-stroke survivors typically try to \textit{``remember what they learn in a therapeutic session and replicate it themselves''} (PS 5). Whether a post-stroke survivor has difficulty with recalling and starting an exercise, post-stroke survivors desire a way that can brief the program of exercises and introduce a new exercise to make them more self-confident and engaged in rehabilitation. 

\textit{``When I do a prescribed exercise, I just keep recalling and doing it to memorize’’}. (PS 4). \textit{``I have memorized exercises well. But sometimes I still need to know how to use a tool or perform an exercise (...) I would rather not perform it if I do not know how to do it”} (PS 1).

\textit{``I sometimes have trouble with remembering a list of exercises from a therapist. If I do remember it later, I choose an exercise depending on my location in the house”} (PS 3). \textit{``As I rely on my memory to conduct exercises, I end up randomly performing a mixed set of exercises from a therapist and online videos, and feel less organized and engaged”} (PS 2).

\subsubsection{Correct Execution of Exercises}\label{sect:hri-study-design-int-patients-monitor}
Even if post-stroke survivors periodically receive therapeutic sessions, they still become confused about whether they perform an exercise correctly. They strive to exercise correctly as much as possible. However, they sometimes end up performing incorrectly without the supervision of a therapist.

 \textit{``Not doing right can be harmful to my progress. I try to keep as closely as possible to what I learned during therapy sessions”} (PS 2). \textit{``But I always do not know if my motion is correct or not’’} (PS 1). 
\textit{``The way I perform exercise is not 100\% correct. At times, I end up doing it wrong”} (PS 3). \textit{``We can never guarantee whether we do exercises correctly by ourselves unless there is someone like a therapist”} (PS 4).

\subsubsection{Strategies to Modulate Difficulty and Pain}\label{sect:hri-study-design-int-patients-adapt}
During rehabilitation, post-stroke survivors with limited functional abilities \textit{``cannot complete a motion fully”} (PS 5) and inevitably \textit{``experience various pains”} (PS 3). Even though they desire to keep practicing exercises by themselves for improvement, these pains and fears of performing alone usually prevent their active participation in rehabilitation \cite{o2019physical}. Some post-stroke survivors prefer to stop doing self-paced sessions due to their concerns about any undesirable, negative consequences. The other two post-stroke survivors (PS 3 and PS 5) preferred to keep performing a simpler exercise while paying attention to avoid injury and pain.

\textit{``We need to keep trying an exercise to improve even if it seems challenging”} (PS 4). 
\textit{``However, even if I prefer not to stop, sometimes I cannot do it alone without the support and supervision of a therapist. I hope to have something that can keep helping me like a therapist”} (PS 2).

\textit{``I need to be more cautious when I do alone”} (PS 3).
\textit{`If exercises are too difficult, I feel more tired and pain. I give up an exercise to relieve pain with my self-taught strategy (e.g. riding a static bicycle), because I am worried about a dangerous situation, another lesion or not making progress. If I have some supervisors that could help to check the correct execution of an exercise and guide how to fix any incorrect motion, I would try again even if I have little tiredness or pain’’} (PS 1).

\subsection{Lack of Systematic Management, Records on Progress}\label{sect:hri-study-design-int-patients-conclude}
Similar to the planning of self-paced rehabilitation sessions, post-stroke survivors do not have any systematic management or records to track their progress. Instead, they simply \textit{``count their repetitions by memory’’} (PS 2) and primarily rely on their \textit{``subjective feelings to understand and check any minor improvement’’} (PS 5) on their functional abilities. As such improvement takes a long time and is barely noticeable from one session to the other session, patients have lack of information on tracking their progress. 

\textit{``I do not keep track of my progress”} (PS 3). \textit{``I just try to check up with the physician whether I have any progress. After a long period of rehabilitation, I start to grasp a bottle and bring it to my mouth, which I use to check my progress”} (PS 1).

\textit{``I feel mentally how much I could move my arm forward more than yesterday to track my progress”} (PS 4). \textit{``I have problems with memorizing and understanding my achievement and minor progress”} (PS 2). 

\subsection{Probing Ideas of Technological Support}
All post-stroke survivors showed positive expectations of technological support. They also expressed willingness to learn and adopt new technology, but highlighted the importance of providing desirable features and being easy to use. 


\textit{``Even after stroke, I have learned how to use a computer, so we can learn and use new technologies for our benefits. However, these technologies should be accessible to use and keep us engaged through providing adequate assistance on rehabilitation''}  (PS 1) while \textit{``moving around my home”} (PS 3).

Post-stroke survivors provided high-level ideas and suggestions on technological support to compensate for the lack of therapist assistance during their self-paced rehabilitation.
\textit{``A person can be forgetful''} (PS 3), so \textit{``scheduling the day and time of a session on a calendar and keeping a person attentive can be good''} (PS 4). \textit{``I want a tool that would make me more organized on schedules of self-directed rehabilitation''} (PS 5).

\textit{``It would be nice to introduce the goals of a session, what needs to be achieved. For instance, if a system shows the list of exercises and how to do on a screen, I could easily follow what needs to be done”} (PS 4). \textit{``As I do not have to put my efforts to memorize and recall, I could become relieved and have focused on what I need to do”} (PS 5).
\textit{``Presenting a program and order of exercises and introducing how to do it like a therapist can be very helpful and make me reliable”} (PS 1) and \textit{``motivated to work”} (PS 2).  

Post-stroke survivors also desire a system that can \textit{``provide instructional, corrective feedback on their performance''} (PS 5) and \textit{``modulate the difficulty based on their fatigue''} (PS 1) or \textit{``pain”} (PS 3). \textit{``As doing an exercise incorrectly can be more harmful, I would like to receive information whether I do an exercise correctly or not''} (PS 1).
\textit{``I wonder if a system can use a video camera to record how I do and correct me when I do an exercise in a wrong way through verbal feedback''} (PS 2). 
\textit{``Pain from attempting to achieve a full-motion is one of the worst''} (PS 3). \textit{``When I cannot achieve it fully, I want it to adjust the difficulty and encourage me to make more effort''} (PS 2).
\textit{``doing at least half of the goal would be better than just quitting''} (PS 4). 

In addition, post-stroke survivors need a system that could keep track of and inform their progress to enhance self-awareness and engagement. 
\textit{``I do not know how, but I want a system that can analyze and provide detailed information, the progress of my performance on multiple aspects of motor skills''} (PS 5). \textit{``so that I can check and understand my recovery progress every morning or week. This information would make me more engaged with my rehabilitation''} (PS 2).

\section{Practices of Therapists to Guide Rehabilitation Sessions}\label{sect:hri-study-design-int-therapists}

After conducting initial interviews with post-stroke survivors to understand their needs, we also interviewed therapists to understand their procedures and strategies for operating rehabilitation sessions and interacting with post-stroke survivors. The summarized findings are described in Table \ref{tab:findings-therapists-practices-coach}.

\begin{table*}[htp!]
\centering
\caption{Findings of Therapists' Practices to Operate a Session and Generate Feedback}
\label{tab:findings-therapists-practices-coach}
\resizebox{\textwidth}{!}{%
\begin{tabular}{ccl} \toprule
\multicolumn{2}{c}{\textbf{Findings}} &
  \multicolumn{1}{c}{\textbf{Details}} \\ \midrule
\multicolumn{2}{c}{Procedures of an interactive session} &
  \begin{tabular}[c]{@{}l@{}}1) Introduction: a) brief greeting and b) instructing a motion\\ 2) Run a session: \\  $\;\;\;$ a) monitor an exercise\\ $\;\;\;$ b) provide feedback for improvement \& encouragement\\ $\;\;\;$ c) understand the status to adapt a session\\ $\;\;\;$ e) summarize patient's performance\end{tabular} \\ \midrule
\multicolumn{1}{l}{\multirow{3}{*}{\begin{tabular}[c]{@{}l@{}}Interaction - \\ Communications \&\\ Feedback\end{tabular}}} &
  Objectives &
  1) Instructional, 2) Motivational, 3) Building a Relationship, 4) Clarifying the Status \\
\multicolumn{1}{l}{} &
  Types &
  1) Visual, 2) Verbal, 3) Physical \\
\multicolumn{1}{l}{} &
  Timing &
  \begin{tabular}[c]{@{}l@{}}1) Before, 2) After, and 3) During a motion\end{tabular} \\ \midrule
\multirow{2}{*}{Considerations} &
  Physical &
  \begin{tabular}[c]{@{}l@{}}1) Whether a motion is complete, smooth, involves unnecessary, compensatory joints\\ 2) Whether a patient seems tired\end{tabular} \\
 &
  Emotional &
  1) Motivation, 2) Perceived Level of Difficulty \\ \bottomrule
\end{tabular}%
}
\end{table*}

\subsection{Interactive Rehabilitation Session}
Rehabilitation therapy is important to address the injuries or illnesses that refrain a person's abilities to move and conduct activities of daily living \cite{o2019physical}. Results from the initial interviews with therapists made clear that rehabilitation therapy requires the active participation of post-stroke survivors and interactions between therapists and post-stroke survivors. 
During a rehabilitation therapy session, therapists oversee the treatments through tailoring the goal of a post-stroke survivor, instructing and encouraging the survivor for the correct execution of treatments. At the same time, post-stroke survivors also require proactive commitment and communication to clarify and share their statuses that might not be easily noticeable from therapists.

\textit{``During a session, I try to understand the post-stroke survivor's status and help the survivor's engagement in rehabilitation for recovery. It is not just instructing an exercise (...) I aim to understand various factors of a post-stroke survivor from what I see and communicate. Depending on the post-stroke survivor's status and feedback, I make sure to provide an adequate intervention''} (TP 1)

\subsection{Overall Procedures of an Interactive Session}
We found that therapists have common procedures to manage a session with post-stroke survivors. Depending on the status of a post-stroke survivor, therapists typically arrange one to three sessions per week with a post-stroke survivor. At the beginning of a session, therapists \textit{``provide a brief greeting''} (TP 3) and \textit{``describe and instruct what a post-stroke survivor would work during a session''} (TP 2). When a post-stroke survivor performs an exercise, therapists \textit{``observe how an individual performs an exercise''} (TP 3) and \textit{``provide positive encouragement and corrective feedback on how a post-stroke survivor can improve''} (TP 2). In addition, therapists also engaged in conversation to understand the internal status of a post-stroke survivor and adapt a session accordingly. \textit{``Depending on post-stroke survivor's feedback, I determine whether to push more or not''} (TP 1). At the end of a session, therapists summarize how well a post-stroke survivor performs and ask a post-stroke survivor to work alone on a particular aspect of a motion until the next session (e.g. reducing the usage of post-stroke survivor's shoulder joint while raising the hand to the mouth).

\subsection{Interactions of Therapists for a Session}\label{sect:hri-study-design-int-tps-communication}
Our findings on interviews with therapists provide detailed insights on how therapists interact with post-stroke survivors (e.g. communications and generating feedback). 

\subsubsection{Objectives}
Interactions between therapists and post-stroke survivors can be broadly classified into the following four objectives: 1) instructing, 2) motivating a post-stroke survivor 3) building a relationship, and 4) clarification on the status of a post-stroke survivor. 

As the correct execution of an exercise is critical to improve post-stroke survivor's functional abilities, therapists \textit{``explain how a post-stroke survivor performs an exercise and which aspects a post-stroke survivor should be mindful''} (TP 3). At the same time, therapists also provide positive encouragement to participate in rehabilitation, \textit{``as most post-stroke survivors would have difficulty with completing an exercise''} (TP 2). In addition, therapists \textit{``engage in small talk with post-stroke survivors to build a relationship''} (TP 1) and \textit{``ask the clarification on post-stroke survivor's status that cannot be easily determined by observation''} (TP 3)

\subsubsection{Visual, Verbal, \& Physical Interactions}
Therapists interact with a post-stroke survivor through the following three modalities: visual instructions with gestures, verbal communications, and physical contact. 

For visual instructions, therapists \textit{``perform an exercise or replicate a post-stroke survivor's incorrect motion''} (TP 2) to instruct the correct execution of an exercise. They also provide verbal descriptions, encouragement, and feedback to complement their explanations with gestures. When a post-stroke survivor has low strength and difficulty with completing an exercise, therapists provide \textit{``physical support to give cues and achieve a movement''} (TP 1). In addition, therapists engage in a small talk with the post-stroke survivor to build a relationship over sessions, such as \textit{``how are you doing?''} (TP 3). They also ask for clarification on what they speculate about the status of a post-stroke survivor: \textit{``this exercise seems too challenging for you. would you like to continue more with a lower target position?''} (TP 1). 

\subsubsection{Timing}
Therapists mentioned that typically, they aim to engage in conversations with a post-stroke survivor or provide feedback immediately when a particular situation occurs. However, they sometimes refrain from generating immediate and repetitive feedback to avoid making a post-stroke survivor more frustrated. They mostly engaged with a post-stroke survivor before and after a patient completes an exercise. \textit{``I usually give feedback immediately right after completing an exercise or when a post-stroke survivor performs an incorrect motion''} (TP 2). \textit{``But, it depends. If I just continuously ask a post-stroke survivor to avoid compensation, it would make him more frustrated (...) When a post-stroke survivor performs an incorrect motion with compensated joints, I would rather speak differently and suggest taking a rest while''} (TP 1).

\subsection{Understanding Physical and Emotional Status for Tailored Rehabilitation}\label{sect:hri-study-design-int-tps-status}
Therapists highlighted the importance of understanding the status of a post-stroke survivor to provide an adequate session and recommended considering physical and emotional factors to guide a session. As post-stroke survivors have diverse physical and emotional conditions, therapists aim to build a mutual relationship with post-stroke survivors over sessions to better understand their status and tailor a session accordingly. 

\textit{``There are a lot of reasons why post-stroke survivors cannot do an exercise properly (...) we have to understand the status of a post-stroke survivor (...) how a post-stroke survivor moves, whether the post-stroke survivor is tired or has any pains (...) so that we determine what kinds of feedback would work the best to have good performance''} (TP 1).

\textit{``I aim to make an exercise to be challenging, but also avoid asking the thing that a post-stroke survivor cannot do. But this could be very different for each person''} (TP 3). 

\textit{``For example, I have one post-stroke survivor, who thinks that she needs to suffer pains and push to get improvement. In contrast, I have another post-stroke survivor, who starts complaining and stops immediately if he feels a little bit of pain''} (TP 1).

Our results suggest that understanding the status of a post-stroke survivor can be categorized into physical and emotional aspects. Therapists first observe how a post-stroke survivor performs to understand various physical conditions. For instance, therapists \textit{``check whether a post-stroke survivor has lack of strength, tremors, tiredness to complete an exercise''} (TP 2). In addition, as performing rehabilitation over an extended period can be challenging for post-stroke survivors, therapists strive to understand the post-stroke survivor's emotional status. For example, therapists \textit{``pay attention to whether a post-stroke survivor feels frustrated and motivated to participate in an exercise and an exercise is challenging enough''} (TP 1).

\subsection{Possibility of Technological Support}\label{sect:hri-study-design-int-tps-techsupport}
All therapists were uncertain about the capabilities of technology to monitor post-stroke survivor's exercise and mentioned a few concerns on technological support for post-stroke survivor's self-directed rehabilitation. Specifically, they considered that a system could \textit{``show a video to instruct an exercise, and provide audio-based encouragement''} (TP 2). However, therapists have doubts on how well technology can understand the correct execution of post-stroke survivor's exercises and the emotional aspects of a post-stroke survivor. \textit{``Can technology observe post-stroke survivor's exercise and understand whether a post-stroke survivor completes an exercise and performs any compensated motions like lifting post-stroke survivor's shoulder? (...) Also, I wonder how a system could determine whether an exercise is challenging enough for a post-stroke survivor} (TP 1). 

Therapists highlighted the necessity of considering both physical and emotional aspects of a post-stroke survivor to provide adequate feedback in the case of developing a system that supports post-stroke survivor's rehabilitation. \textit{``If a system could monitor that a post-stroke survivor cannot complete an exercise, the system should not just repetitively say `do not compensate'. Instead, it should make an adjustment on a task or discuss post-stroke survivor's preference to take a rest''} (TP 1).
\section{{Prototyping} an AI and Robotic Coach for Physical Rehabilitation Therapy}\label{sect:hri-study-design-results-prototype}
\begin{figure}[h]  \includegraphics[width=1.0\columnwidth]{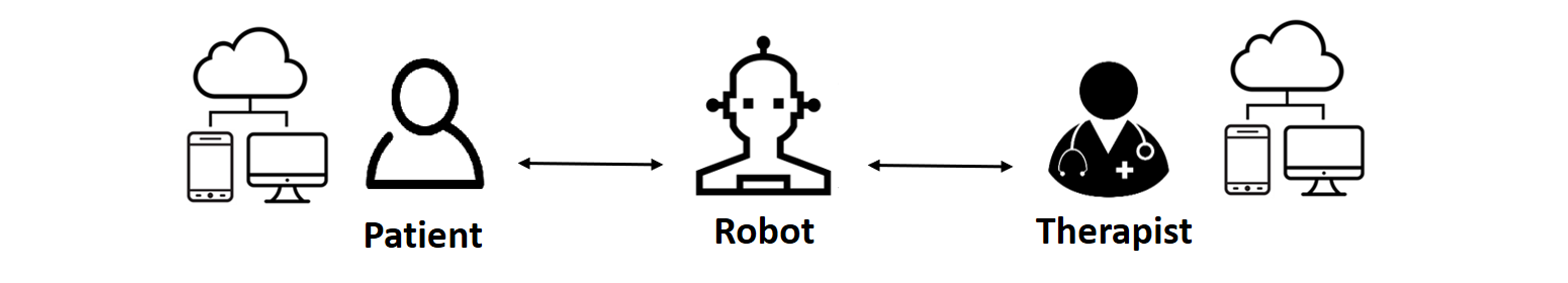}
  \caption{Flow diagram of AI and robotic coach that interacts with a patient and a therapist to support rehabilitation practices}\label{fig:hri-design-system-F0}
\end{figure}

Based on analysis of initial interviews with post-stroke survivors and therapists (Section \ref{sect:hri-study-design-int-patients} and \ref{sect:hri-study-design-int-therapists}), the research team designed and developed an AI and robotic coach (Figure \ref{fig:hri-design-system-F0}). Specifically, the team first utilized the needs of post-stroke survivors to specify the major functionalities of the system. After determining the functionalities, the team also leveraged the practices of therapists and initial thoughts of technological support from post-stroke survivors and therapists to further design low-level specifications on how the system can assist post-stroke survivor's self-directed rehabilitation and develop a high fidelity prototype. 

\subsection{Functionalities}
\begin{figure}[h]{}
\centering
  \includegraphics[width=1.0\columnwidth]{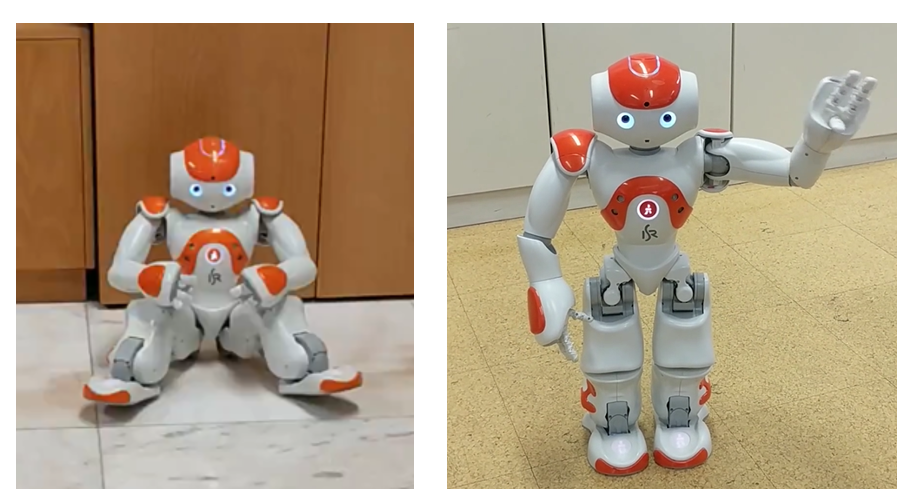}
  \caption{An AI and robotic coach receives the signal of the post-stroke survivor's smartphone at the time of a scheduled session, approaches the post-stroke survivor, and gives salutation to initiate a session.}
  \label{fig:hri-design-system-F2}
\end{figure}

Table \ref{tab:hri-study-design-prototype-overview} describes the challenges and needs of post-stroke survivors during their self-directed rehabilitation and the corresponding functionalities of an AI and robotic coach. These functionalities include F1) planning a session, F2) initiating a session, F3) introducing a session, F4) monitoring and providing corrective feedback, F5) adapting the difficulty of a session, and F6) concluding a session (Figure \ref{fig:hri-design-system-F1},  \ref{fig:hri-design-system-F2},  \ref{fig:hri-design-system-F36}). In the following section, we further describe each functionality of our AI and robotic coach in detail along with its alignment with therapists' practices (Table \ref{tab:exampleinteractions}).

\subsubsection{Planning a Session}
\begin{figure}[h]
\centering 
\begin{subfigure}[t]{\columnwidth}
  \centering
  \includegraphics[width=1.0\columnwidth]{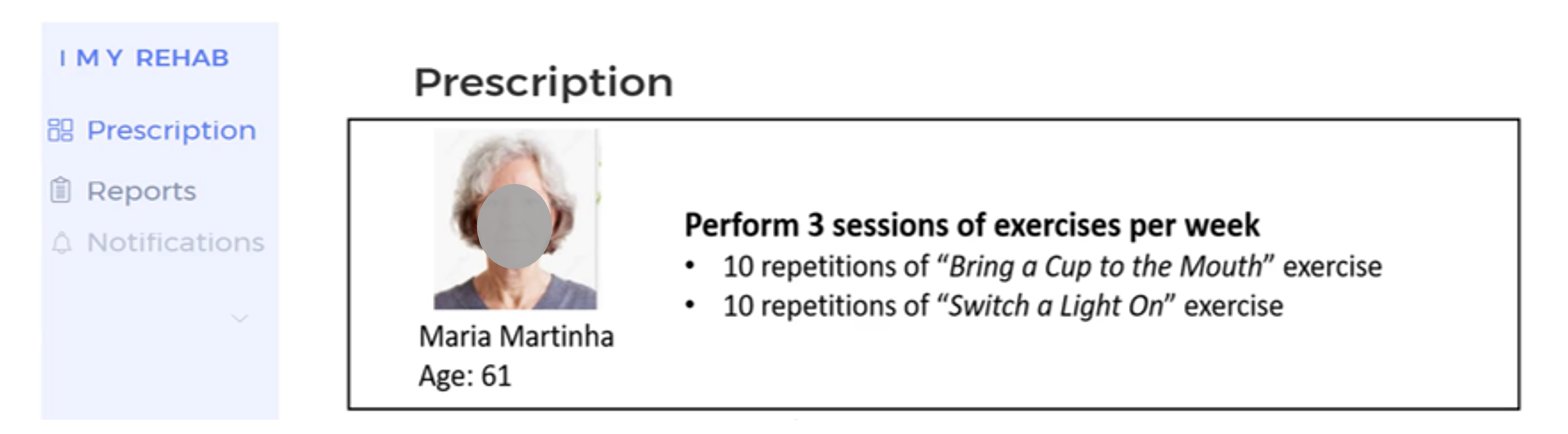}
  \caption{}
  \label{fig:hri-design-system-F1-1}
\end{subfigure}
\begin{subfigure}[t]{\columnwidth}
  \centering
  \includegraphics[width=1.0\columnwidth]{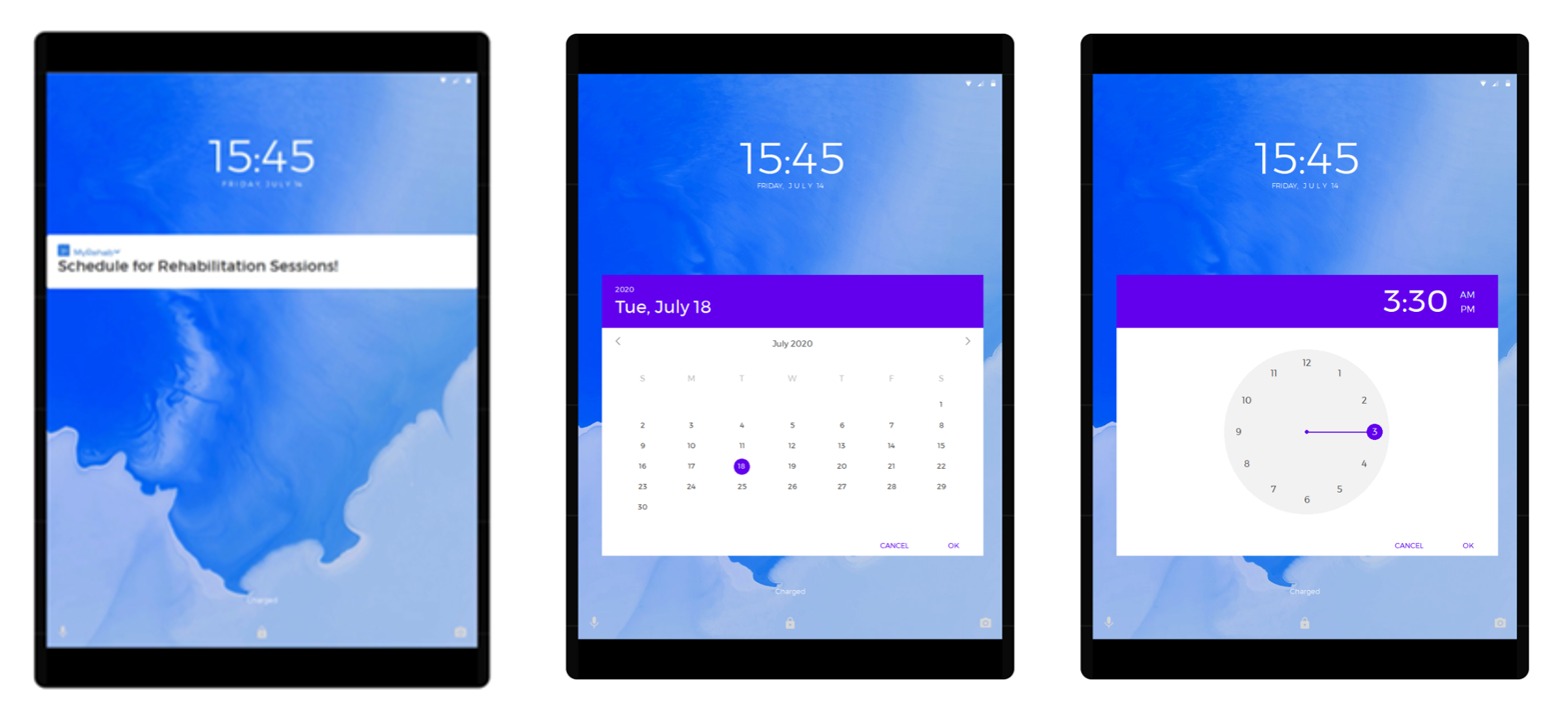}
  \caption{}
  \label{fig:hri-design-system-F1-2}
\end{subfigure}
\caption{(a) Interface for therapists to upload a prescription for post-stroke survivor's self-directed rehabilitation. (b) Interface of a mobile phone for post-stroke survivors to receive a notification and plan a session.}\label{fig:hri-design-system-F1}
\end{figure}

One primary function of the system is to allow post-stroke survivors flexible planning on self-directed rehabilitation. As a smartphone starts being widely adopted by elderly people and explored for health services \cite{boulos2011smartphones,fortney2011re}, the team decided that post-stroke survivors could make a flexible plan of their self-directed rehabilitation through a smartphone instead of relying on spontaneous planning (Section \ref{sect:hri-study-design-int-patients-planning}). When a therapist uploads the prescriptions of exercises for a post-stroke survivor on a web interface (Figure \ref{fig:hri-design-system-F1-1}), the post-stroke survivor would receive a notification on a smartphone to schedule the day and time of a session on the calendar interface (Figure \ref{fig:hri-design-system-F1-2}). The post-stroke survivor would receive another notification on a smartphone at the time and day of the scheduled session.

\subsubsection{Initiation of a Session}
As a simple notification on a smartphone could be dismissed by a post-stroke survivor, the team envisioned that an AI and robotic coach could facilitate the initiation of a self-directed session with the post-stroke survivor. If the post-stroke survivor does not initiate a session even after the scheduled time, an AI and robotic coach could approach the user and engage in a dialogue to greet and ask the user's intention of starting a session (Figure \ref{fig:hri-design-system-F2}).  

\subsubsection{Introduction of a Session}
\begin{figure*}[htp]
\centering
\begin{subfigure}[]{0.9\textwidth}
  \centering
  \includegraphics[width=1.0\columnwidth]{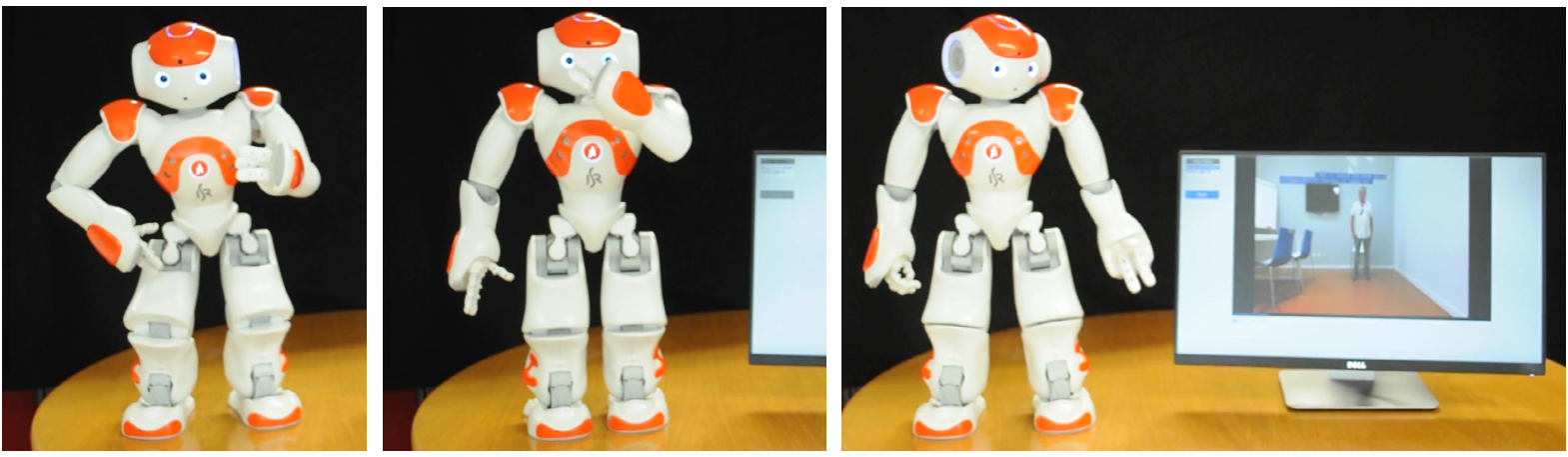}
  \caption{Introduction of a Session}
  \label{fig:hri-design-system-F3}
\end{subfigure}
\begin{subfigure}[]{0.9\textwidth}
  \centering
  \includegraphics[width=1.0\columnwidth]{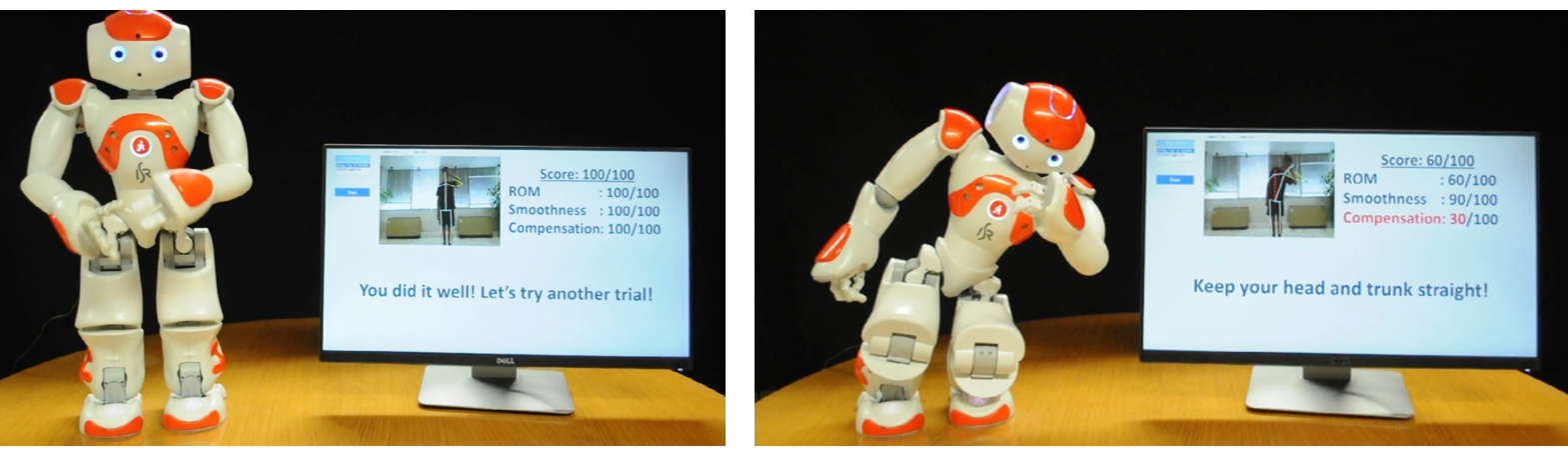}
  \caption{Monitoring and Providing Feedback}
  \label{fig:hri-design-system-F4}
\end{subfigure}
\begin{subfigure}[]{1.0\textwidth}
  \centering
  \includegraphics[width=1.0\columnwidth]{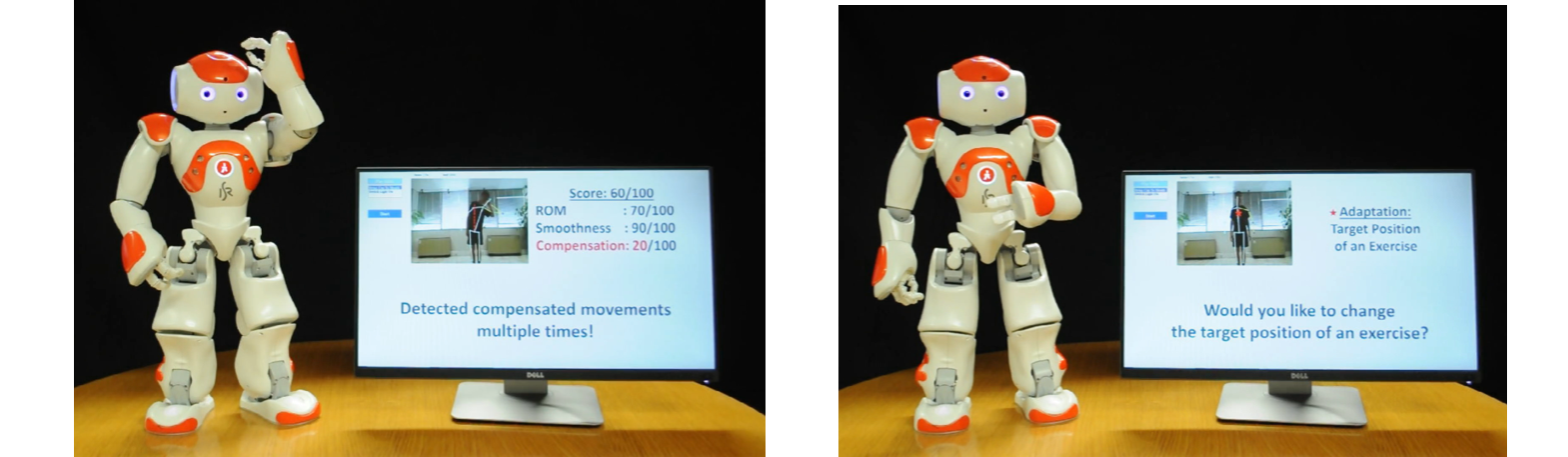}
  \caption{Adapting the Difficulty of a Session}
  \label{fig:hri-design-system-F5}
\end{subfigure}
\begin{subfigure}[]{0.95\textwidth}
\centering
  \includegraphics[width=1.0\columnwidth]{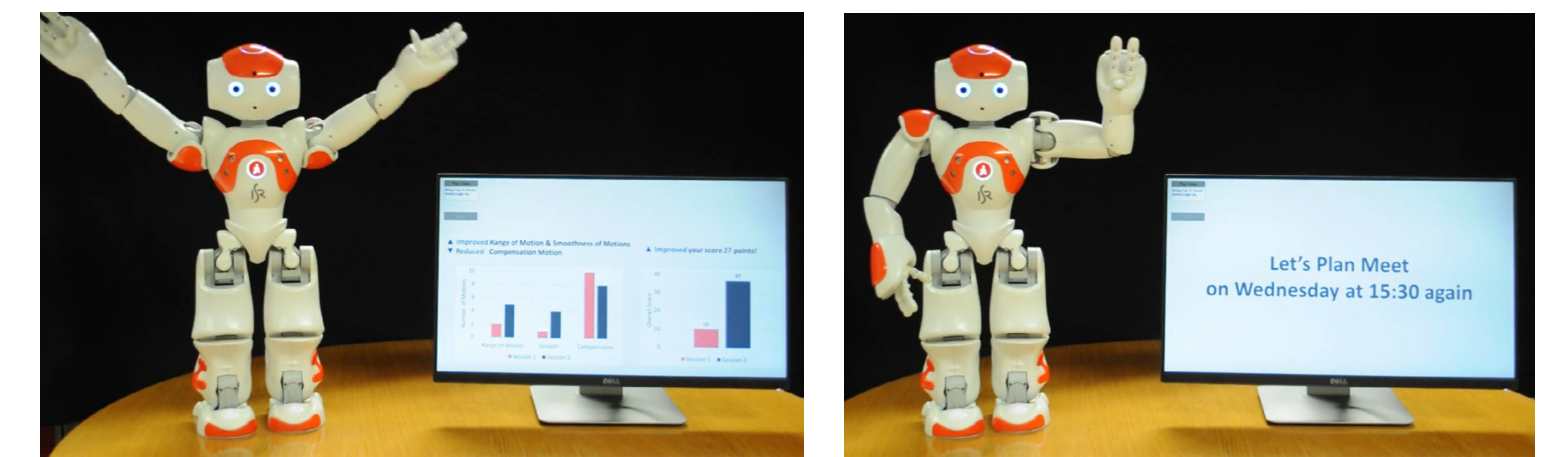}
  \caption{Concluding a Session}
  \label{fig:hri-design-system-F6}
\end{subfigure}
\caption{Functionalities of an AI and Robotic Coach: (a) an AI and robotic coach describes the goal of a session, demonstrates an exercise with gestures, and shows a video on the display. (b) an AI and robotic coach monitors and assesses the exercises of the post-stroke survivor and provides encouragement and corrective feedback with gestures, audios, and visualization. (c) once an AI and robotic coach detects that the post-stroke survivor continuously performs an exercise with compensated joints, the AI and robotic coach communicates with the post-stroke survivor to adjust the difficulty of a session. (d) an AI and robotic coach summarizes the overall performance and progress of the post-stroke survivor, and reminds the next scheduled session.}\label{fig:hri-design-system-F36}
\end{figure*}

The team determined that an AI and robotic coach could assist post-stroke survivors to recall what they have to perform instead of relying on their memory (Section \ref{sect:hri-study-design-int-patients-program}). Specifically, an AI and robotic coach could brief the goal of a session prescribed by a therapist, since presenting a goal is an important factor to increase behavior change  \cite{scobbie2013implementing}. In addition, an AI and robotic coach could demonstrate how to perform an exercise with gestures and visualization on a display (Figure \ref{fig:hri-design-system-F3}). 

\begin{table*}[htp]
\centering
\caption{Example interactions of our AI and robotic coach and their mapping to therapists' practices}
\label{tab:exampleinteractions}
\resizebox{\textwidth}{!}{%
\begin{tabular}{clcll}\toprule
\textbf{Functionalities} &
  \multicolumn{1}{c}{\textbf{Objectives}} &
  \textbf{Types} &
  \multicolumn{1}{c}{\textbf{Timing}} &
  \multicolumn{1}{c}{\textbf{Example Interactions}} \\ \midrule
\multirow{2}{*}{F2. Initiation} &
  \multirow{2}{*}{\begin{tabular}[c]{@{}l@{}}Building a relationship\\ \& Clarifying the status\end{tabular}} &
  Verbal &
  \multirow{2}{*}{Before a motion} &
  \begin{tabular}[c]{@{}l@{}}\textit{``Hello Maria, How are you doing?''},  \textit{`Are you ready to start exercises?''}\end{tabular} \\
 &
   &
  Visual &
   &
  A robot waves a hand \\ \midrule
\multirow{2}{*}{F3. Introduction} &
  \multirow{2}{*}{Instructional} &
  Verbal &
  \multirow{3}{*}{Before a motion} &
  \begin{tabular}[c]{@{}l@{}}\textit{``Today, we will conduct 10 repetitions of bring a cup to the mouth exercise''}\\ \textit{``You need to place your hand to the mouth as if drinking wate''}\end{tabular} \\
 &
   &
  Visual &
   &
  \begin{tabular}[c]{@{}l@{}}Playing a video demonstration on the interface. A robot demonstrates an exercise (Figure \ref{fig:hri-design-system-F3})\end{tabular} \\ \midrule
\multirow{6}{*}{\begin{tabular}[c]{@{}c@{}}F4. Monitoring\\ \& Feedback\end{tabular}} &
  \multirow{4}{*}{Instructional} &
  \multirow{2}{*}{Verbal} &
  During a motion &
  \textit{``Please do not move your trunk to the side''} \\
 &
   &
   &
  After a motion &
  \begin{tabular}[c]{@{}l@{}}\textit{``You did it well''}. \textit{``But I found you have an incomplete, non-smooth, compensatory motion''}\end{tabular} \\
 &
   &
  \multirow{2}{*}{Visual} &
  During a motion &
  Displaying patient's joint positions on the interface \\
 &
   &
   &
  After a motion &
  A robot imitates a patient's incorrect motion \\
 &
  \multirow{2}{*}{Motivational} &
  Verbal &
  \multirow{2}{*}{After a motion} &
  \begin{tabular}[c]{@{}l@{}}\textit{``You can do better''}, \textit{``You did a great job''}, \textit{``Let's keep going''}, \textit{``You have only x more trials''}\end{tabular} \\
 &
   &
  Visual &
   &
  A robot provides encouraging gestures, such as clapping (Figure \ref{fig:hri-design-system-F4}) \\ \midrule
\multirow{2}{*}{\begin{tabular}[c]{@{}c@{}}F5. Adapting \\ Difficulty\end{tabular}} &
  \multirow{2}{*}{Clarifying the status} &
  Verbal &
  \multirow{2}{*}{After a motion} &
  \begin{tabular}[c]{@{}l@{}}\textit{``I found that you performed compensation multiple times''}\\ \textit{``Would you like to change the target position and continue or stop exercising?''}\end{tabular} \\
 &
   &
  Visual &
   &
  Displaying verbal feedback in texts on the interface along with a robot gesture (Figure \ref{fig:hri-design-system-F5}) \\ \midrule
\multirow{6}{*}{F6. Concluding} &
  \multirow{2}{*}{Instructional} &
  Verbal &
  \multirow{6}{*}{After a motion} &
  \begin{tabular}[c]{@{}l@{}}\textit{``Compared to the previous session, you reduced incomplete, non-smooth, compensatory motion''}\end{tabular} \\
 &
   &
  Visual &
   &
  Displaying a comparative graph on patient's performance \\
 &
  \multirow{2}{*}{Motivational} &
  Verbal &
   &
  \textit{``Great job, you completed all trials''} \\
 &
   &
  Visual &
   &
  A robot puts both hands in the air and waves its hand (Figure \ref{fig:hri-design-system-F6})\\
 &
  \multirow{2}{*}{Building a relationship} &
  Verbal &
   &
  \textit{``Let's meet up again next on the next appointment schedule (Wed, 15:30pm)''} \\
 &
   &
  Visual &
   &
  \begin{tabular}[c]{@{}l@{}}A robot waves its hands, displaying a next appointment schedule on the interface (Figure \ref{fig:hri-design-system-F6})\end{tabular} \\ \bottomrule
\end{tabular} 
} 
\end{table*}

\subsubsection{Monitoring an Exercise and Providing Feedback}
The team specified that AI and robotic coaches should be able to automatically monitor and assess the quality of patient's exercises to provide corrective feedback. We hypothesize that corrective feedback of an AI and robotic coach has the potential to address post-stroke survivor's confusion on how to correctly perform an exercise themselves (Section \ref{sect:hri-study-design-int-patients-monitor}). Following the practices of therapists (Section \ref{sect:hri-study-design-int-therapists}), an AI and robotic coach could visualize how a post-stroke survivor performs an exercise on a display, provide audio-based encouragement and corrective feedback, and instruct how to correctly perform an exercise with gestures (Figure \ref{fig:hri-design-system-F4}). Specifically, an AI and robotic coach can analyze the quality of motion in terms of the range of motion (i.e. how closely a post-stroke survivor achieves the target position), smoothness of a motion, and compensation (i.e. whether a post-stroke survivor utilizes an unnecessary joint to make a motion) \cite{sullivan2011fugl,lee2020exploratory}. When a post-stroke survivor completes an exercise correctly, an AI and robotic coach provides positive encouragement with the gesture of clapping. If the post-stroke survivor incorrectly performs an exercise (e.g. learning trunk to the side), an AI and robotic coach can instruct the user by replicating and visualizing a user's incorrect motion and providing audio corrective feedback for improvement (e.g. \textit{``keep your trunk straight''}) \cite{lee2021interactive}. Although therapists also make physical contact with post-stroke survivors to provide corrective feedback on a joint position, the team decided to focus on the social interaction of an AI and robotic coach due to safety concerns \cite{gorgey2018robotic}.

\subsubsection{Adjusting the Difficulty of a Session}\label{sect:functions_adaptation}
An AI and robotic coach should consider both physical and emotional aspects of a post-stroke survivor (Section \ref{sect:hri-study-design-int-tps-techsupport}) to provide an adequate session for a post-stroke survivor, who inevitably experiences pains and feel unmotivated or afraid of any undesirable consequences (Section \ref{sect:hri-study-design-int-patients-adapt}). Physical aspects include the completion of a motion and a potential source of pain for post-stroke survivors (e.g. the occurrence of excessive, repetitive compensated motions, feeling tiredness). Emotional aspects include the level of frustration or motivation but are not limited to those. When a post-stroke survivor repetitively performs a compensated motion, an AI and robotic coach should engage in a dialogue with the post-stroke survivor to understand the user status and recommend adjusting the target position of an exercise if necessary or taking a rest (Figure \ref{fig:hri-design-system-F5}).  

\subsubsection{Concluding a Session}
Even if post-stroke survivors with greater self-awareness of their deficits are more likely to participate in rehabilitation \cite{fleming1995self}, they do not have any systematic management or records on their progress (Section \ref{sect:hri-study-design-int-patients-conclude}). Thus, the team included the functionality of an AI and robotic coach to keep track of post-stroke survivor's progress, summarize how the performance of a post-stroke survivor changes over sessions, and remind about the next scheduled session (Figure \ref{fig:hri-design-system-F6}).

\subsection{High-Fidelity Prototype}
\begin{figure}[h] 
\centering
\includegraphics[width=0.5\columnwidth]{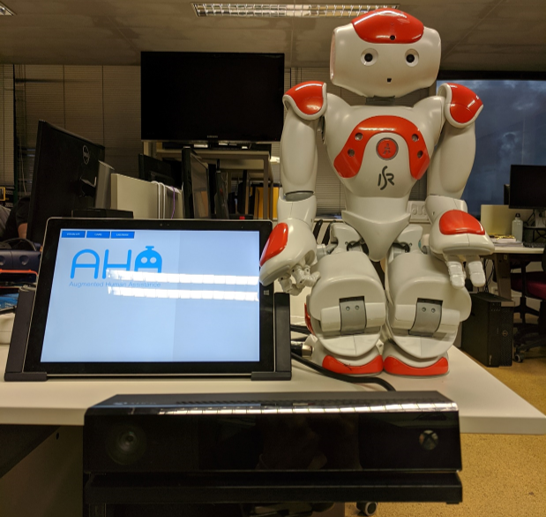}
  \caption{System prototype that includes a Kinect sensor to track a patient's exercise and a social robot, NAO and a visualization interface to provide gesture-based, audio, and visual feedback on patient's performance}\label{fig:system}
\end{figure}

Based on the low-level specifications of an AI and robotic coach, the team developed a high-fidelity prototype. The overall system is composed of a social robot, a Kinect sensor, and a visualization interface (Figure \ref{fig:system}).  

For the functionality of \textit{``F1: Planning''}, the team utilized a wireframe tool to demonstrate a potential web interface for therapists and a smartphone application for post-stroke survivors. As the main focus of this work is to evaluate and collect early opinions on the potential of an AI and robotic coach, we did not implement the entire pipeline of a server and communication between a web interface for therapists, an AI and robotic coach, and a smartphone application for post-stroke survivors. 

For the interactions with post-stroke survivors, the team decided to explore the feasibility of a social robot, following the prior work that describes the benefit of physical embodiment to improve the user's engagement in exercises \cite{fasola2013socially} (Section \ref{sect:hri-study-design-related-app-patients}). Specifically, the team used an NAO robot, because it supports competitive hardware capabilities with cost reduction \cite{gouaillier2008nao} to implement our specific functionalities (Table \ref{tab:hri-study-design-prototype-overview}), but also a user-friendly software development environment. This NAO robot is compact and lightweight with a height of 0.57m and a weight of 4.5 kg. For the functionality of \textit{``F2: Initiating''}, the NAO supports smooth bi-ped walking to approach a post-stroke survivor. For other functionalities, the NAO robot also supports high degrees of freedom (DOF) to provide gesture-based feedback (e.g. replicating a prescribed exercise or a patient's motion). 
The team utilized the NAO SDK \cite{pot2009choregraphe} to implement an NAO program that controls the gestures of the NAO. 

For monitoring an exercise, the team utilized a Kinect v2 sensor, which has the benefit of being non-invasive than a wearable sensor (Section \ref{sect:hri-study-design-related-motion-track}), but also recording the images and video of patient's exercises (Figure \ref{fig:hri-design-system-F4}). The team developed a monitoring program that tracks patient's exercises with Kinect SDK and presents images of patient's exercises with overlaid skeletons. In addition, this program automatically assesses the quality of post-stroke survivor's exercise \cite{lee2020towards} and provide visual feedback on the tablet screen and audio feedback using Google TTS libraries \cite{kepuska2017comparing} and the tablet speaker. These NAO and monitoring programs operate on a tablet and are connected through a socket programming. Even if the NAO SDK supports to use a built-in video camera and Text To Speech (TTS) libraries, we did not utilize them due to its unstable IP connection.

After developing a high-fidelity prototype, the team recorded a narrated video demonstration of the prototype, which would serve as a foundation to receive opinions from therapists and post-stroke survivors. Before conducting the evaluation study, we reviewed the narrated videos with two therapists (TPs with check marks in the review column of Table \ref{tab:hri-study-design-tps}). Overall, they could easily understand how the system could interact with a post-stroke survivor. One minor comment from them was to make the speed of narrations in a video slightly slower, so that post-stroke survivors could follow the contents of a video better. This comment was addressed before conducting the evaluation study with new therapists and post-stroke survivors.

\section{Evaluation}

\begin{figure*}[t]
\centering 
\begin{subfigure}[t]{1.0\textwidth}
\centering
  \includegraphics[width=1.0\columnwidth]{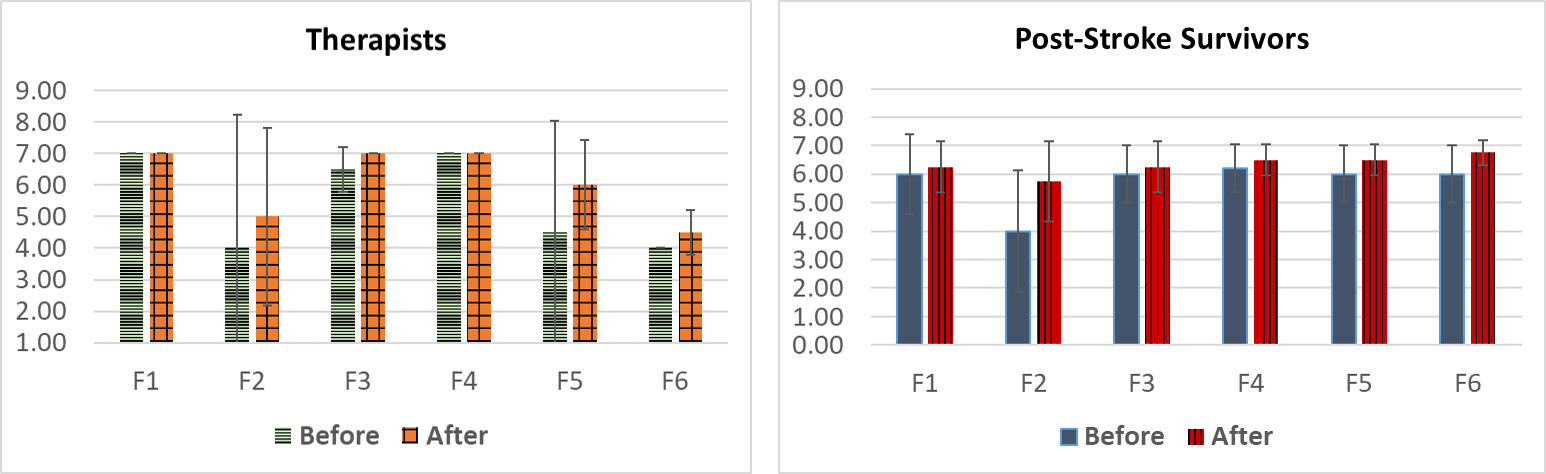}
  \caption{Expectation of Technological Supports}
  \label{fig:hri-design-results-evaluation-expect}
\end{subfigure}\par\vspace{0.2cm} 
\begin{subfigure}[t]{1.0\textwidth}
\centering
  \includegraphics[width=1.0\columnwidth]{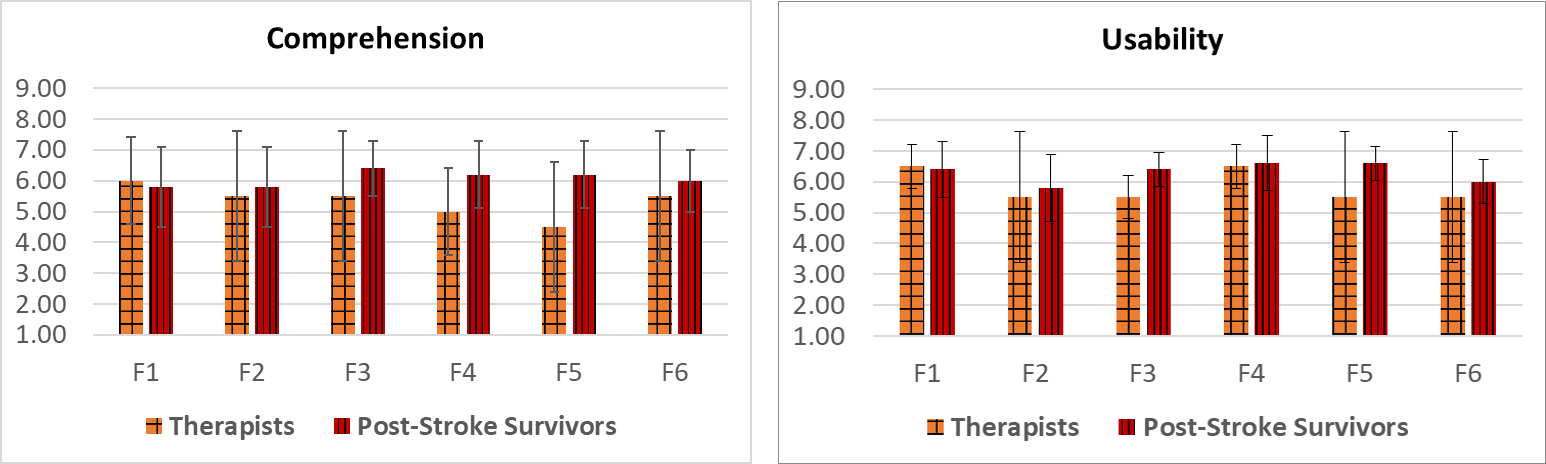}
  \caption{Comprehension and Usability Scores of the System}
  \label{fig:hri-design-results-evaluation-usability}
\end{subfigure}
\caption{Results of Questionnaires. (a) the expectation of technological support before and after showing the videos of the system (b) scores on comprehension and usability scores of the system from therapists and post-stroke survivors.  }\label{fig:hri-design-results-evaluation-all}
\end{figure*}

\cite{dkfjdfj}
After developing a high-fidelity prototype, we conducted additional interviews with therapists and post-stroke survivors to understand their opinions on technological supports for self-directed rehabilitation. Figure \ref{fig:hri-design-results-evaluation-all} summarizes the quantitative responses to questionnaires (i.e. expectation before and after showing the video of a prototype, comprehension, and usability of a prototype) from therapists and post-stroke survivors. 

{
Overall, both therapists and post-stroke survivors have shown positive responses (i.e. above 3.5 out of 7) on the six major functionalities of our AI and robotic coach with all aspects of evaluation metrics (i.e. expectation, comprehension, and usability). Specifically, post-stroke survivors described the potential benefits of our AI and robotic coach that can support systematic management to better coordinate self-directed rehabilitation sessions, instruct an exercise convincingly and clearly to make a post-stroke survivor more secure and motivated to do an exercise. However, they have also expressed several concerns: 1) possible difficulty of interaction for post-stroke survivors with cognitive impairment, 2) portability, and 3) costs.}

\subsection{{Expectation}}
Both therapists and post-stroke survivors showed positive expectations of technological supports before and after showing the video of our prototype. After seeing the video, they developed more concrete ideas on how technological supports can improve post-stroke survivor’s self-directed rehabilitation, which leads to more positive expectations of technological supports. Specifically, after reviewing the videos, therapists provided higher average expectation score with lower standard deviation from 5.50 $\pm$ 1.41 to 6.08 $\pm$ 0.82.
Similarly, post-stroke survivors also provided higher average expectation score with lower standard deviation from 5.70 $\pm$ 1.23 to 6.33 $\pm$ 0.79. 

\textit{``Initially, I have some doubts about how a system could help post-stroke survivors, but after watching videos, I have a better idea of the system. It was very cute and interesting to have this robot. Some post-stroke survivors or even other populations would enjoy this concept''} (TP 3).

\textit{``Yes!! (sound with excitement with the possibility). we cannot always have a therapist. I want to have a machine that can coordinate a session, check my performance, adjust things accordingly, and provide encouragement to make more effort''} (PS 2).

\subsection{{Comprehension and Usability}}
In addition, both therapists and post-stroke survivors considered that interactions and functionalities of a prototype in the video were comprehensive and useful. For comprehension, therapists provide an average score of 5.33 $\pm$ 0.51, and post-stroke survivors rated an average score of 6.07 $\pm$ 0.24. For usability, therapists provide an average score of 5.83 $\pm$ 0.51, and post-stroke survivors provide an average score of 6.30 $\pm$ 0.32.

Therapists clarified that they do not have any issues with a prototype in terms of comprehension and usability aspects as they \textit{``do not have any cognitive limitation''} (TP 4). However, they mentioned that even the same interactions with a prototype \textit{``might become difficult to be understood by some post-stroke survivors with cognitive limitation''}. 

When we evaluated the video of the prototype with post-stroke survivors, one post-stroke survivor (PS 5), who sometimes requires assistance on his daily living activities, could not initially understand the image of a person with overlaid skeletons, so the interviewer had to explain verbally and replay the confusion part of the video. Except for PS 5, most post-stroke survivors considered that they could interact with the presented prototype easily and considered such a prototype as very helpful to support self-directed rehabilitation. 
\textit{``I think I would be able to use it easily, and understand the system better over interactions. I really like it and would like to give the system the highest score as this system could be very useful}'' (PS 3).

\subsection{Planning a Session}
Post-stroke survivors considered that planning a session with a smartphone application and receiving notification would be \textit{``beneficial to check the list of prescribed exercises”} (PS 1), and get reminded of what they have to achieve. \textit{``I can arrange sessions with the flexibility to my availability”} (PS 2) and \textit{``It would become easier and faster to coordinate my sessions as I always carry my smartphone''} (PS 4).  

 There are two post-stroke survivors (PS 3 \& 5), who do not have a smartphone and do not know how to use it. However, these patients still expressed positively about the functionality of our system and their willingness to learn and use it. 
\textit{``As I forget, this would be good to just mark my availability and remind me of the `homework’ to work on it (..) But I do not have a smartphone. If I had a phone, I would learn how to use it as this function can help me remember''} (PS 3).

\subsection{Initiation of a Session}
Therapists described that the initiation of a session should be done by a post-stroke survivor if the post-stroke survivor is independent to perform an exercise. However, even if some post-stroke survivors do not have an issue of memorizing a session, they find the value of receiving technical support to initiate a session in a certain situation when they accidentally forget a session or feel demotivated.

\textit{``I never had any robots, but it would not be a bad idea. I changed to have more positive expectations on the usage of a robot”} (PS 5). \textit{``For my case, it is not necessary all the time. However, it seems easy to interact with this robot. One day, I may be busy and forget to do exercises. This robot could remind me to start a session''} (PS 2).

\textit{``This can benefit more for stroke survivors with a memory issue or low motivation to remind and avoid slacking off. With the robot that says ‘Hey, wake up! Let’s move for the exercises’, I can get more enthusiastic''} (PS 3).

\textit{``Sometimes, I do not feel like doing anything, just staying on a chair. As doing nothing leads to nowhere, I would like to have this foreman to come and provide alert and encouragement to wake up''} (PS 1). 

\textit{``The robot approaching to start a session is an interesting addition. I could have more incentive and willingness to initiate my rehabilitation sessions with the robot''} (PS 4).

\subsection{Introduction of a Session}
Similar to the initiation of a session, patients did not have a severe problem of recalling the program of exercises. However, they still appreciate systematic support for better management and learning a new strategy or exercise.

\textit{``It’s a nice idea to brief the goal of a session and demonstrate an exercise with a robot and video. I would better remember what I need to do''} (PS 1).
\textit{``This robot did explain very well and clearly. As I forget easily, it would be very useful to better understand what I need to do''} (PS 3). 

\textit{``Instead of relying on my memory, this could help keep tracking what I have to do''} (PS 5) and
\textit{``recall and organize better the program of an exercise''} (PS 4). \textit{``This would make me motivated to keep working on the goal of a session''} (PS 2).

\subsection{Monitoring an Exercise and Providing Corrective Feedback}
Post-stroke survivors considered that performing an exercise correctly is critical:
\textit{``We would like to know our body positions and know right postures to improve our functional ability''} (PS 1). \textit{``When I do it myself, I do not have a clear perception on how my movement was done''} (PS 4).  \textit{``I like to see the image of my motion that displays my body joints. I can see what I am doing, and understand whether I perform correctly or not''} (PS 3).

Post-stroke survivors reiterated that the functionality of monitoring and providing feedback is \textit{``quite useful and important as a guide, coach of rehabilitation (…) the system presents various types of information on how well I perform comprehensively and correct any incorrect way. It was easy to understand''} (PS 1). \textit{``It is very useful to help people to do exercises correctly''} (PS 4). \textit{``Having this corrective feedback would be beneficial as incorrect movements can occur for any patient. It’s similar to having physiotherapy while being at home''} (PS 2). 

Among various types of information from the system, PS 4, who has a hearing impairment, expressed that \textit{``verbal feedback is useful but more difficult for me to follow, but I can follow and understand from our body images and the gestures of the robot''}. In contrast, PS 5 found that \textit{``verbal feedback is the most important and valuable information''}. Other post-stroke survivors mentioned that \textit{``all feedback would work better and complement each other when they present together''} (PS 4). PS 3 mentioned that she could \textit{``better understand how I need to adjust and correct my motion''} through observing the image display of post-stroke survivor’s motions, listening to audio feedback, and checking additional gestures from an AI and robotic coach.

\subsection{Adjusting the Difficulty of a Session}\label{sect:hri-study-design-evaluation-F5}
Post-stroke survivors inevitably experienced difficulty with completing a prescribed exercise, and felt insecure about performing self-paced sessions. They expressed the benefits of having a system that communicates to understand their status and adjust the goal of a session for better engagement in self-directed rehabilitation. 

\textit{``As I have difficulty with doing certain movements, I like how the system understands my excessive effort without success to provide necessary adjustment''} (PS 2). 
\textit{``It is very beneficial and positive to keep trying''} (PS 4). 

\textit{``I am cautious and afraid of doing certain exercises without supervision. (…) I like how the system makes an adaptation that starts with a simpler exercise and gradually updates''} (PS 3).
\textit{``I would like to have this functionality that communicates to understand my status and supports having a different goal to reach''} (PS 5).  

\textit{``I am convinced how the robot adjusts the difficulty of an exercise. It is very helpful in a way that I can gradually participate in an exercise while preventing me from performing it incorrectly. Before I mentioned that I would be hesitant to try again when an exercise is challenging. But, I would try again with this system''} (PS 1). \textit{``This gradual adaptation makes more secure and confident to do an exercise''} (PS 3).

\subsection{Concluding a Session}
Post-stroke survivors described that concluding a session with a summary of their progress would assist them to have better self-awareness of their progress, but also engagement and motivation in their rehabilitation. 

\textit{``I never had this formal tracking of my progress when I did rehabilitation. I liked the features of this system. Presenting the progress with graphs is easy to understand, and could assist people to have a better notion of how their efforts make difference''} (PS 4).

\textit{``It is great to inform with more clear numerical values on how they make progress over a session. Currently, I cannot observe my progress easily through my subjective feelings, and therapists do not tell me which gains I make. With this richer information on my progress, I would have a better mindset and try my best to improve my progress next time''} (PS 1).

\textit{``Having a system that can keep track of everything I do is great''} (PS 2). \textit{``I could check what I have already achieved before and after each session”} (PS 3). \textit{``after some days or weeks, the system could show me 'look you achieve this, so later we aim for a higher goal' to make me more engaged and keep my progress informed to my therapist''} (PS 2).

\subsection{Diverse Styles and Preferences on Interactions with AI and Robotic Coaches}
Post-stroke survivors described various styles and preferences to interact with an AI and robotic coach. For instance, some post-stroke survivors preferred to have less autonomy of an AI and robotic coach: 

\textit{``I would be able to interact with the robot when it comes to reminding me to start a session, but as I could recall well, I would like to have more autonomy on starting a session''} (PS 2).  Furthermore, the PS 2 does not think that small talks with an AI and robotic coach \textit{``is necessary''} and prefer to only receive rehabilitation-related information.

In contrast, other post-stroke survivors preferred to have more active, autonomous behaviors of an AI and robotic coach. 
\textit{``I liked that the robot would look for me to call me for a session. If not, why would I have it?''} (PS 3).
\textit{``A person needs to start when it is time''} (PS 4). \textit{``when a person is not active, a robot could come and ask to start a session as if being our boss on rehabilitation session (..) as I notice robot’s intention of reminding is to not skip a session, I would take it seriously and try to follow''} (PS 1). 
\textit{``having small talks with a system can be more natural and receptive. Not just staying there and staring at me''} (PS 1). 

We also found that an individual, post-stroke survivor can have different preferences of engagement and autonomy on each functionality of the system. For instance, PS 4 did not desire small talk with an AI and robotic coach, but he found it interesting and willing to start a session at the request of an approached, AI and robotic coach. 

Similarly, we also observed diverse preferences of post-stroke survivors about how they would receive different types of feedback on an exercise. 
PS 1 mentioned that presenting different types of information (e.g. visualization, audio, and gestures from the robot) \textit{``does not interfere with each other rather they reinforce}. In contrast, P2 commented on his preference \textit{``to interpret information separately''}. 

\subsection{Other Considerations: Portability and Cost}
Even if our study focuses on the interaction with an AI and robotic coach, post-stroke survivors also provided comments on other factors, such as the portability, size, and cost of a robot. For instance, P1 mentioned that he has to travel often between two cities and \textit{``carrying it from one place to the other would be a disadvantage''}. P3 provided an additional comment that \textit{``I found the robot is useful. However, having this robot, another big object at home, could be bothersome''}. 
In terms of the cost, PS 5 described that he could save up costs of travel (e.g. taking a taxi) to visit a rehabilitation center by having a system that supports self-directed rehabilitation. At the same time, PS 2 described that \textit{``I want to pursue self-paced rehabilitation sessions without spending more money (e.g. buying a robot)''}.

\section{Discussion}
Throughout the iterative involvement of therapists and post-stroke survivors, we explored the feasibility of an AI and robotic coach to assist self-directed rehabilitation in an effective and acceptable way. In the following sections, we discussed the potential implications of our exploratory study and considerations for better deployment of AI and robotic coaches in practice.

\subsection{Early Involvement of Stakeholders for Better Acceptance}
Our work has demonstrated that the involvements of stakeholders (e.g. both therapists and post-stroke survivors) were critical to deriving an AI and robotic coach in a more acceptable way. Specifically, they assisted to produce a broad set of functionalities to address post-stroke survivor's challenges during the entire process of self-directed rehabilitation. In addition, they provided actionable and detailed specifications on design ideas on how an AI and robotic coach can interact with and guide post-stroke survivor's self-directed rehabilitation while following therapists' practices. As these design ideas were grounded by stakeholders' practices and experiences, our high-level prototype received positive expectations and high comprehension and usability scores from both therapists and post-stroke survivors. 

Our evaluation results showed that expectations of therapists and post-stroke survivors were mismatched in some cases. For instance, therapists considered that it is preferable for post-stroke survivors to make an initiative on self-directed rehabilitation if they have capability to do it independently. However, as rehabilitation requires an engagement over an extended period \cite{o2019physical}, post-stroke survivors could become demotivated and unwilling to make an initiation. Post-stroke survivors considered that initiation and interaction with an AI and robotic coach could make them more motivated. Thus, we recommend that researchers involve not only therapists, who can support eliciting clinically grounded designs, but also the end-user (e.g. post-stroke survivors), who will interact with the system to enhance the acceptance of a system. 

\subsection{Micro, Function-Level Personalization}
Our results  provide another insight on personalization for an AI and robotic coach. Specifically, this work suggests personalization of an AI and robotic coach should be considered at micro, functional-level for better acceptance. For instance, when we explored several functionalities of an AI and robotic coach, each post-stroke survivor has unique preferences and styles of interactions with a system. In addition, a post-stroke survivor expressed two contradictory styles of interactions with a system. For instance, PS4 did not desire an active engagement (e.g. small talk) with a system. In contrast, PS4 was positive of an AI and robotic coach approaching him to assist the initiation of a session. If we just apply a uniform interaction style of a system over functionalities (e.g. identifying the personality of a user to specify a single interaction style \cite{tapus2008user}), such a system could lead to lower user acceptance in some functionalities. Thus, we recommend enabling a system to personalize a micro, functional-level based on the user's status and needs.

\subsection{Options of Re-Explanations and Addressing a System Failure}
In addition, our study suggests that it is necessary to provide an option of re-explanations and addressing a failure. Throughout the evaluation study, all therapists and most post-stroke survivors easily comprehended the interactions presented in the video demonstration of a prototype. However, there was one post-stroke survivor (PS 5), who became confused about the overlaid skeleton on an image. Further clarification that the overlaid dots, skeletons on an image is to facilitate checking body positions of a post-stroke survivor during an exercise was necessary from the interviewer to PS 5. We consider that such requests of re-explanations could occur at any aspect of functionalities from a system. In addition, a system failure could also occur during a real-world deployment. Thus, our finding recommends that a system should be able to re-explain and address a failure for better communication and acceptance from the user in practice.  

\subsection{Interactive Dialogues to Explain and Understand the Status of a User}
Our study also draws attention to the potential value of an interactive dialogue to clarify the status and functionalities of a system, but also understands subjective user's status. An interactive dialogue provides a natural way to explain the status and functionalities of a system to user \cite{fitter1979towards,devault2014simsensei} and has the benefit of providing a more positive and acceptable experience for the user \cite{saini2005benefits}. 
Moreover, our results showed that interactive dialogue responses of a system can augment the capability of a system to understand subjective user status (Section \ref{sect:hri-study-design-evaluation-F5}). For instance, as therapists highlighted to consider both physical and emotional aspects of a user to provide an adequate, personalized session, we derived the design specifications to detect whether an exercise is considered too challenging or not and adjust the difficulty of a session (Section \ref{sect:functions_adaptation}). Even if significant recent work aims to make a fully automated machine learning (ML) model to understand the emotional states of a user with a complex algorithm, it is still challenging to achieve a decent performance \cite{ng2015deep}. Thus, we explored an alternative approach, which does not rely on an ML model to detect whether an exercise is challenging or not. Instead, we implemented a system to check the occurrence of a compensated motion (e.g. leaning trunk to the side) \cite{lee2020towards}. Once a compensated motion is consecutively detected multiple times, we utilized such detection to trigger an interactive dialogue response with a user and confirm the user's subjective status. 
According to the evaluation with post-stroke survivors about the functionality of adjusting the difficulty of a session, all post-stroke survivors provided positive feedback. Specifically, they understood why adjustment discussion is initiated. In addition, they appreciated that such gradual adaptation will make them engaged in rehabilitation more securely and confidently.
Thus, this work discusses the value of making a system interactive to support a better understanding of the status of a system and a user instead of just creating a fully automated approach.

\subsection{Potential Impact and Limitations}
AI and robotics coaches are increasingly employed in the domain of healthcare \cite{tsui2011want,riek2017healthcare} (e.g. monitoring well-being related or rehabilitation exercises \cite{fasola2013socially,mataric2007socially,lee2020towards,lee2021human}). However, the evaluation of these systems is limited to a specific function (e.g. monitoring an exercise). The deployment of these systems is still challenging. Our results showed that early involvements of stakeholders was critical to gain insights from them and derive more effective and acceptable AI and robotic coaches. Our study demonstrated the potential of an AI and robotic coach to assist post-stroke survivor's engagement in self-directed rehabilitation through six major functionalities (e.g. planning, initiation, introduction, adjustment, and conclusion of a session). We also discussed a few considerations (e.g. interactive techniques, micro functional-level personalization, portability, and cost of a system) to bring us closer to realize these AI and robotic coaches in practice. In addition, as an AI and robotic coach requires to collect videos, audios, and functional status of patients, it is also important to provide an adequate means of controlling data to preserve their privacy.

Our study is limited to recruit post-stroke survivors, who do not have any cognitive impairment to conduct interviews, and have a small sample size, which does not represent all therapists and post-stroke survivors. However, such a small sample size is not unusual among similar studies \cite{azenkot2016enabling,feingold2020social}. It is important to further explore how to make these systems accessible for people with cognitive impairment.

In addition, as our study is intended to inform preliminary exploration on effective and acceptable interaction with AI and robotic coaches to assist self-directed rehabilitation therapy, therapists and post-stroke survivors had limited interactions with an AI and robotic coach through the video demonstration of a prototype. It is necessary to make a system prototype robust to operate in real-world. In addition to the focus of this study, AI and robotic systems to assist post-stroke survivor's self-directed rehabilitation therapy, it is important to explore how to make AI and robotic systems more acceptable in the perspective of therapists to administrate post-stroke survivor's rehabilitation \cite{lee2021human}. Further real-world study on a more complex task is required to better understand the applicability and generalization of an AI and robotic coach in various domains.

\section{Conclusion}
In this work, we have explored the feasibility of AI and robotic coaches to assist self-directed stroke rehabilitation through the iterative involvement of both therapists and post-stroke survivors. Specifically, we co-designed, developed, and evaluated an AI and robotic coach that assists the overall process of self-directed rehabilitation instead of focusing on a particular function, procedure of rehabilitation (e.g. monitoring an exercise). While deploying these systems in practice is still challenging, this work discusses the potential of an AI and robotic coach to support a simple task for the user without cognitive impairment. In addition, we recommend key considerations for better deployment of AI and robotic coaches: an involvement of stakeholders in an early design phase, micro, functional-level personalization, and interactive dialogues to communicate the status of a system and a user.

\begin{acknowledgements}
The authors thank Carolina Vieira for the assistance with running and coding the interviews with participants, and all the participants in this study for their time and valuable inputs. This work is partially supported by the National Science Foundation (NSF) under grant number CNS-1518865. Additional support was provided by the IntelligentCare project (LISBOA-01-0247-FEDER-045948), the FCT LARSyS funding 2020-2023 (UIDB/50009/2020), the FCT project HAVATAR (PTDC/EEI-ROB/1155/2020), and
the Singapore Ministry of Education (MOE) Academic Research
Fund (AcRF) Tier 1 grant.
\end{acknowledgements}

%
\section*{Conflict of interest}
The authors have no conflicts of interest to declare that are relevant to the content of this article.
%

\section*{Data Availability Statement}
The datasets generated during and/or analyzed during the current study are available from the corresponding author on reasonable request.


\bibliographystyle{spmpsci}
\bibliography{main}

\begin{thebibliography}{10}
\providecommand{\url}[1]{{#1}}
\providecommand{\urlprefix}{URL }
\expandafter\ifx\csname urlstyle\endcsname\relax
  \providecommand{\doi}[1]{DOI~\discretionary{}{}{}#1}\else
  \providecommand{\doi}{DOI~\discretionary{}{}{}\begingroup
  \urlstyle{rm}\Url}\fi

\bibitem{alankus2010towards}
Alankus, G., Lazar, A., May, M., Kelleher, C.: Towards customizable games for
  stroke rehabilitation.
\newblock In: Proceedings of the SIGCHI conference on human factors in
  computing systems, pp. 2113--2122 (2010)

\bibitem{argent2018patient}
Argent, R., Daly, A., Caulfield, B.: Patient involvement with home-based
  exercise programs: can connected health interventions influence adherence?
\newblock JMIR mHealth and uHealth \textbf{6}(3), e47 (2018)

\bibitem{azenkot2016enabling}
Azenkot, S., Feng, C., Cakmak, M.: Enabling building service robots to guide
  blind people a participatory design approach.
\newblock In: 2016 11th ACM/IEEE International Conference on Human-Robot
  Interaction (HRI), pp. 3--10. IEEE (2016)

\bibitem{baillie2019challenges}
Baillie, L., Breazeal, C., Denman, P., Foster, M.E., Fischer, K., Cauchard,
  J.R.: The challenges of working on social robots that collaborate with
  people.
\newblock In: Extended Abstracts of the 2019 CHI Conference on Human Factors in
  Computing Systems, pp. 1--7 (2019)

\bibitem{beer2012domesticated}
Beer, J.M., Smarr, C.A., Chen, T.L., Prakash, A., Mitzner, T.L., Kemp, C.C.,
  Rogers, W.A.: The domesticated robot: design guidelines for assisting older
  adults to age in place.
\newblock In: Proceedings of the seventh annual ACM/IEEE international
  conference on Human-Robot Interaction, pp. 335--342 (2012)

\bibitem{boulos2011smartphones}
Boulos, M.N.K., Wheeler, S., Tavares, C., Jones, R.: How smartphones are
  changing the face of mobile and participatory healthcare: an overview, with
  example from ecaalyx.
\newblock Biomedical engineering online \textbf{10}(1), 24 (2011)

\bibitem{do2016movement}
do~Carmo Vilas-Boas, M., Cunha, J.P.S.: Movement quantification in neurological
  diseases: methods and applications.
\newblock IEEE reviews in biomedical engineering \textbf{9}, 15--31 (2016)

\bibitem{czaja2006factors}
Czaja, S.J., Charness, N., Fisk, A.D., Hertzog, C., Nair, S.N., Rogers, W.A.,
  Sharit, J.: Factors predicting the use of technology: Findings from the
  center for research and education on aging and technology enhancement
  (create).
\newblock Psychology and aging \textbf{21}(2), 333 (2006)

\bibitem{devault2014simsensei}
DeVault, D., Artstein, R., Benn, G., Dey, T., Fast, E., Gainer, A., Georgila,
  K., Gratch, J., Hartholt, A., Lhommet, M., et~al.: Simsensei kiosk: A virtual
  human interviewer for healthcare decision support.
\newblock In: Proceedings of the 2014 international conference on Autonomous
  agents and multi-agent systems, pp. 1061--1068 (2014)

\bibitem{ekstam2007functioning}
Ekstam, L., Uppgard, B., Von~Koch, L., Tham, K.: Functioning in everyday life
  after stroke: a longitudinal study of elderly people receiving rehabilitation
  at home.
\newblock Scandinavian journal of caring sciences \textbf{21}(4), 434--446
  (2007)

\bibitem{fasola2013socially}
Fasola, J., Matari{\'c}, M.J.: A socially assistive robot exercise coach for
  the elderly.
\newblock Journal of Human-Robot Interaction \textbf{2}(2), 3--32 (2013)

\bibitem{feingold2020social}
Feingold~Polak, R., Tzedek, S.L.: Social robot for rehabilitation: Expert
  clinicians and post-stroke patients' evaluation following a long-term
  intervention.
\newblock In: Proceedings of the 2020 ACM/IEEE International Conference on
  Human-Robot Interaction, pp. 151--160 (2020)

\bibitem{fitter1979towards}
Fitter, M.: Towards more “natural” interactive systems.
\newblock International Journal of Man-Machine Studies \textbf{11}(3), 339--350
  (1979)

\bibitem{fleming1995self}
Fleming, J., Strong, J.: Self-awareness of deficits following acquired brain
  injury: Considerations for rehabilitation.
\newblock British Journal of Occupational Therapy \textbf{58}(2), 55--60 (1995)

\bibitem{fortney2011re}
Fortney, J.C., Burgess, J.F., Bosworth, H.B., Booth, B.M., Kaboli, P.J.: A
  re-conceptualization of access for 21st century healthcare.
\newblock Journal of general internal medicine \textbf{26}(2), 639 (2011)

\bibitem{gale2013using}
Gale, N.K., Heath, G., Cameron, E., Rashid, S., Redwood, S.: Using the
  framework method for the analysis of qualitative data in multi-disciplinary
  health research.
\newblock BMC medical research methodology \textbf{13}(1), 117 (2013)

\bibitem{gockley2006encouraging}
Gockley, R., Matari{\'C}, M.J.: Encouraging physical therapy compliance with a
  hands-off mobile robot.
\newblock In: Proceedings of the 1st ACM SIGCHI/SIGART conference on
  Human-robot interaction, pp. 150--155 (2006)

\bibitem{gorgey2018robotic}
Gorgey, A.S.: Robotic exoskeletons: The current pros and cons.
\newblock World journal of orthopedics \textbf{9}(9), 112 (2018)

\bibitem{gouaillier2008nao}
Gouaillier, D., Hugel, V., Blazevic, P., Kilner, C., Monceaux, J., Lafourcade,
  P., Marnier, B., Serre, J., Maisonnier, B.: The nao humanoid: a combination
  of performance and affordability.
\newblock CoRR abs/0807.3223  (2008)

\bibitem{kepuska2017comparing}
K{\"e}puska, V., Bohouta, G.: Comparing speech recognition systems (microsoft
  api, google api and cmu sphinx).
\newblock Int. J. Eng. Res. Appl \textbf{7}(03), 20--24 (2017)

\bibitem{lee2021interactive}
Lee, M.H.: Interactive hybrid intelligence systems for human-ai/robot
  collaboration: Improving the practices of physical stroke rehabilitation.
\newblock Ph.D. thesis, Carnegie Mellon University (2021)

\bibitem{lee2019learning}
Lee, M.H., Siewiorek, D.P., Smailagic, A., Bernadino, A., et~al.: Learning to
  assess the quality of stroke rehabilitation exercises.
\newblock In: Proceedings of the 24th International Conference on Intelligent
  User Interfaces, pp. 218--228. ACM (2019)

\bibitem{lee2020exploratory}
Lee, M.H., Siewiorek, D.P., Smailagic, A., Bernardino, A., Berm{\'u}dez~i
  Badia, S.: An exploratory study on techniques for quantitative assessment of
  stroke rehabilitation exercises.
\newblock In: Proceedings of the 28th ACM Conference on User Modeling,
  Adaptation and Personalization, pp. 303--307 (2020)

\bibitem{lee2021human}
Lee, M.H., Siewiorek, D.P., Smailagic, A., Bernardino, A., Berm{\'u}dez~i
  Badia, S.: A human-ai collaborative approach for clinical decision making on
  rehabilitation assessment.
\newblock In: Proceedings of the 2021 CHI Conference on Human Factors in
  Computing Systems, pp. 1--14 (2021)

\bibitem{lee2020towards}
Lee, M.H., Siewiorek, D.P., Smailagic, A., Bernardino, A., Badia, S.B.: Towards
  personalized interaction and corrective feedback of a socially assistive
  robot for post-stroke rehabilitation therapy.
\newblock In: 2020 29th IEEE International Conference on Robot and Human
  Interactive Communication (RO-MAN), pp. 1366--1373. IEEE (2020)

\bibitem{lo2012exoskeleton}
Lo, H.S., Xie, S.Q.: Exoskeleton robots for upper-limb rehabilitation: State of
  the art and future prospects.
\newblock Medical engineering \& physics \textbf{34}(3), 261--268 (2012)

\bibitem{long2003rehabilitation}
Long, A.F., Kneafsey, R., Ryan, J.: Rehabilitation practice: challenges to
  effective team working.
\newblock International journal of nursing studies \textbf{40}(6), 663--673
  (2003)

\bibitem{van2016rehabilitation}
Van~der Loos, H.M., Reinkensmeyer, D.J., Guglielmelli, E.: Rehabilitation and
  health care robotics.
\newblock In: Springer handbook of robotics, pp. 1685--1728. Springer (2016)

\bibitem{loureiro2011advances}
Loureiro, R.C., Harwin, W.S., Nagai, K., Johnson, M.: Advances in upper limb
  stroke rehabilitation: a technology push.
\newblock Medical \& biological engineering \& computing \textbf{49}(10), 1103
  (2011)

\bibitem{mataric2007socially}
Matari{\'c}, M.J., Eriksson, J., Feil-Seifer, D.J., Winstein, C.J.: Socially
  assistive robotics for post-stroke rehabilitation.
\newblock Journal of NeuroEngineering and Rehabilitation \textbf{4}(1), 5
  (2007)

\bibitem{ng2015deep}
Ng, H.W., Nguyen, V.D., Vonikakis, V., Winkler, S.: Deep learning for emotion
  recognition on small datasets using transfer learning.
\newblock In: Proceedings of the 2015 ACM on international conference on
  multimodal interaction, pp. 443--449 (2015)

\bibitem{o2019physical}
O'Sullivan, S.B., Schmitz, T.J., Fulk, G.: Physical rehabilitation.
\newblock FA Davis (2019)

\bibitem{peek2016interventions}
Peek, K., Sanson-Fisher, R., Mackenzie, L., Carey, M.: Interventions to aid
  patient adherence to physiotherapist prescribed self-management strategies: a
  systematic review.
\newblock Physiotherapy \textbf{102}(2), 127--135 (2016)

\bibitem{pot2009choregraphe}
Pot, E., Monceaux, J., Gelin, R., Maisonnier, B.: Choregraphe: a graphical tool
  for humanoid robot programming.
\newblock In: RO-MAN 2009-The 18th IEEE International Symposium on Robot and
  Human Interactive Communication, pp. 46--51. IEEE (2009)

\bibitem{pripfl2016results}
Pripfl, J., K{\"o}rtner, T., Batko-Klein, D., Hebesberger, D., Weninger, M.,
  Gisinger, C., Frennert, S., Eftring, H., Antona, M., Adami, I., et~al.:
  Results of a real world trial with a mobile social service robot for older
  adults.
\newblock In: 2016 11th ACM/IEEE International Conference on Human-Robot
  Interaction (HRI), pp. 497--498. IEEE (2016)

\bibitem{rensink2009task}
Rensink, M., Schuurmans, M., Lindeman, E., Hafsteinsdottir, T.: Task-oriented
  training in rehabilitation after stroke: systematic review.
\newblock Journal of advanced nursing \textbf{65}(4), 737--754 (2009)

\bibitem{riek2017healthcare}
Riek, L.D.: Healthcare robotics.
\newblock Communications of the ACM \textbf{60}(11), 68--78 (2017)

\bibitem{saini2005benefits}
Saini, P., De~Ruyter, B., Markopoulos, P., Van~Breemen, A.: Benefits of social
  intelligence in home dialogue systems.
\newblock In: IFIP Conference on Human-Computer Interaction, pp. 510--521.
  Springer (2005)

\bibitem{scobbie2013implementing}
Scobbie, L., McLean, D., Dixon, D., Duncan, E., Wyke, S.: Implementing a
  framework for goal setting in community based stroke rehabilitation: a
  process evaluation.
\newblock BMC health services research \textbf{13}(1), 190 (2013)

\bibitem{subramanian2007virtual}
Subramanian, S., Knaut, L.A., Beaudoin, C., McFadyen, B.J., Feldman, A.G.,
  Levin, M.F.: Virtual reality environments for post-stroke arm rehabilitation.
\newblock Journal of neuroengineering and rehabilitation \textbf{4}(1), 20
  (2007)

\bibitem{sullivan2011fugl}
Sullivan, K.J., Tilson, J.K., Cen, S.Y., Rose, D.K., Hershberg, J., Correa, A.,
  Gallichio, J., McLeod, M., Moore, C., Wu, S.S., et~al.: Fugl-meyer assessment
  of sensorimotor function after stroke: standardized training procedure for
  clinical practice and clinical trials.
\newblock Stroke \textbf{42}(2), 427--432 (2011)

\bibitem{swift2015effects}
Swift-Spong, K., Short, E., Wade, E., Matari{\'c}, M.J.: Effects of comparative
  feedback from a socially assistive robot on self-efficacy in post-stroke
  rehabilitation.
\newblock In: 2015 IEEE International Conference on Rehabilitation Robotics
  (ICORR), pp. 764--769. IEEE (2015)

\bibitem{tanguy2016computational}
Tanguy, P., R{\'e}my-N{\'e}ris, O., et~al.: Computational architecture of a
  robot coach for physical exercises in kinaesthetic rehabilitation.
\newblock In: 2016 25th IEEE International Symposium on Robot and Human
  Interactive Communication (RO-MAN), pp. 1138--1143. IEEE (2016)

\bibitem{tapus2008user}
Tapus, A., {\c{T}}{\u{a}}pu{\c{s}}, C., Matari{\'c}, M.J.: User—robot
  personality matching and assistive robot behavior adaptation for post-stroke
  rehabilitation therapy.
\newblock Intelligent Service Robotics \textbf{1}(2), 169 (2008)

\bibitem{thomason2013rehabilitation}
Thomason, P., Graham, H.K.: Rehabilitation of children with cerebral.
\newblock Rehabilitation in Movement Disorders p. 203 (2013)

\bibitem{tsui2011want}
Tsui, K.M., Kim, D.J., Behal, A., Kontak, D., Yanco, H.A.: “i want that”:
  Human-in-the-loop control of a wheelchair-mounted robotic arm.
\newblock Applied Bionics and Biomechanics \textbf{8}(1), 127--147 (2011)

\bibitem{van2004impact}
Van~Peppen, R.P., Kwakkel, G., Wood-Dauphinee, S., Hendriks, H.J., Van~der
  Wees, P.J., Dekker, J.: The impact of physical therapy on functional outcomes
  after stroke: what's the evidence?
\newblock Clinical rehabilitation \textbf{18}(8), 833--862 (2004)

\bibitem{wilson2020challenges}
Wilson, J.R., Tickle-Degnen, L., Scheutz, M.: Challenges in designing a fully
  autonomous socially assistive robot for people with parkinson’s disease.
\newblock ACM Transactions on Human-Robot Interaction (THRI) \textbf{9}(3),
  1--31 (2020)

\bibitem{winkle2018social}
Winkle, K., Caleb-Solly, P., Turton, A., Bremner, P.: Social robots for
  engagement in rehabilitative therapies: Design implications from a study with
  therapists.
\newblock In: Proceedings of the 2018 acm/ieee international conference on
  human-robot interaction, pp. 289--297 (2018)

\bibitem{zhou2008human}
Zhou, H., Hu, H.: Human motion tracking for rehabilitation—a survey.
\newblock Biomedical Signal Processing and Control \textbf{3}(1), 1--18 (2008)

\end{thebibliography}

%
%

\end{document}